\documentclass[conference]{IEEEtran}
\usepackage{times}

\usepackage[numbers]{natbib}
\usepackage[bookmarks=true]{hyperref}

\usepackage{graphicx}
\usepackage{amsmath,amsfonts}
\usepackage{multicol}
\usepackage{multirow}
\usepackage{booktabs}
\usepackage{makecell}
\usepackage{subfig}
\usepackage{float}

\begin{document}

\title{Learning Autonomous Ultrasound via Latent Task Representation and Robotic Skills Adaptation}

\author{\authorblockN{Xutian Deng\authorrefmark{1},
Junnan Jiang\authorrefmark{2},
Wen Cheng\authorrefmark{3} and
Miao Li\authorrefmark{4}}
\authorblockA{\authorrefmark{1}School of Computer Science, Wuhan University, Wuhan 430072, China\\
Email: \href{mailto:dengxutian@whu.edu.cn}{dengxutian@whu.edu.cn}}
\authorblockA{\authorrefmark{2}School of Power and Mechanical Engineering, Wuhan University, Wuhan 430072, China}
\authorblockA{\authorrefmark{3}Hospital of Wuhan University, Wuhan University, Wuhan 430072, China}
\authorblockA{\authorrefmark{4}School of Microelectronics, Wuhan University, Wuhan 430072, China\\
Email: \href{mailto:miao.li@whu.edu.cn}{miao.li@whu.edu.cn}}
}

\maketitle

\begin{abstract}

As medical ultrasound is becoming a prevailing examination approach nowadays, robotic ultrasound systems can facilitate the scanning process and prevent professional sonographers from repetitive and tedious work. Despite the recent progress, it is still a challenge to enable robots to autonomously accomplish the ultrasound examination, which is largely due to the lack of a proper task representation method, and also an adaptation approach to generalize learned skills across different patients. To solve these problems, we propose the latent task representation and the robotic skills adaptation for autonomous ultrasound in this paper. During the offline stage, the multimodal ultrasound skills are merged and encapsulated into a low-dimensional probability model through a fully self-supervised framework, which takes clinically demonstrated ultrasound images, probe orientations, and contact forces into account. During the online stage, the probability model will select and evaluate the optimal prediction. For unstable singularities, the adaptive optimizer fine-tunes them to near and stable predictions in high-confidence regions. Experimental results show that the proposed approach can generate complex ultrasound strategies for diverse populations and achieve significantly better quantitative results than our previous method.

\end{abstract}

\IEEEpeerreviewmaketitle

\section{Introduction}

Medical ultrasound has been widely employed in routine clinical examinations, which faces two main challenges: 1) long-term training for professional skills \cite{arger2005teaching} and 2) physical injury caused by repetitive work \cite{murphy2000update}. To this end, robotic ultrasound systems have been proposed to overcome these challenges \cite{salcudean2022robot, jiang2023robotic}, which can be further divided into teleoperated and autonomous \cite{li2021overview, ning2023autonomous, wang2023task} ones. Autonomous ultrasound contains three major spectra, human-guided \cite{wang2022full}, vision-based \cite{jiang2020automatic, jiang2023dopus} and policy-based, according to the complexity of strategies. Policy-based systems highlight the learning approaches of general ultrasound skills, e.g. learning from demonstrations \cite{droste2020automatic, deng2021robio, raina2023robotic} and reinforcement learning \cite{ning2021autonomic, li2021autonomous, ning2023inverse}. However, these studies have some weaknesses: 1) Since autonomous ultrasound is a high-dimensional robotic task, some indispensable modalities in latent task representation are ignored, e.g. probe orientations and contact forces. 2) The generalization of learned skills, i.e. the adaptation to patients with different physical conditions, has not been proposed and emphasized.

\begin{figure}[t!]
\centering
\includegraphics[width=1\linewidth]{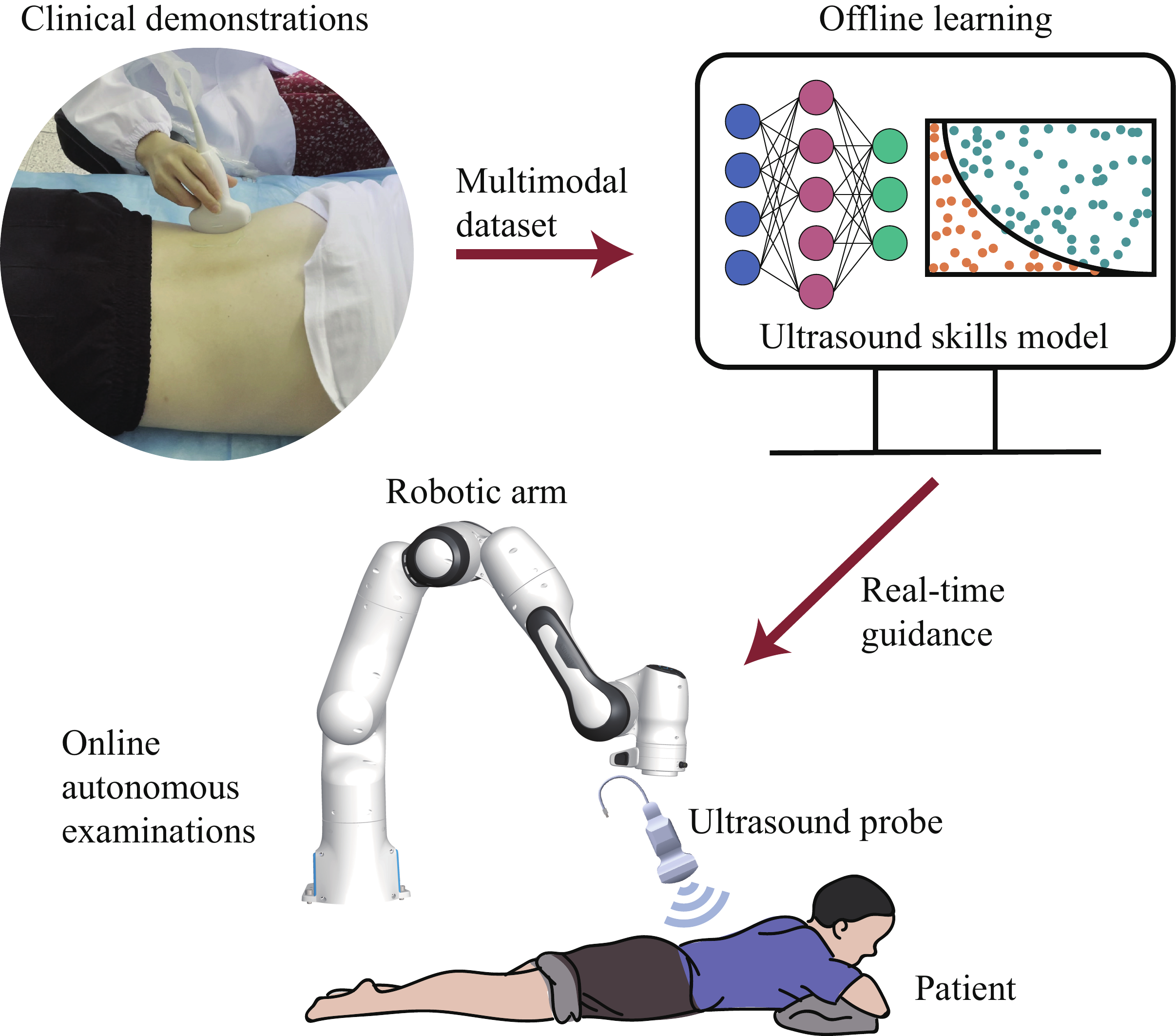}
\caption{\footnotesize The workflow of autonomous ultrasound. First, sonographers' clinical demonstrations for diverse patients are collected. Then, these demonstrated experiences are encapsulated into an ultrasound skills model at the offline stage. Finally, at the online stage, the robotic ultrasound system performs an autonomous examination for a non-previous patient through learned skills.}
\vspace{-6mm}
\label{fig:workflow}
\end{figure}

Following these studies, our work improves the involved modalities with a fully self-supervised framework to perform multimodal fusion and task representation. Besides, we propose and verify an adaptive ultrasound strategy, which abstracts the learned skills into a probability model and accordingly fine-tunes predictions. Fig.~\ref{fig:workflow} shows the workflow of autonomous ultrasound. Our main contribution is twofold:
\begin{itemize}
\item For modeling ultrasound skills, a fully self-supervised latent representation method is proposed, which takes clinically demonstrated ultrasound images, probe orientations, and contact forces into account.
\item For improving the adaptation of learned skills, we propose a probability-based method to evaluate and fine-tune the ultrasound probe’s orientations and contact forces, which enhances the algorithm’s robustness and generalization for patients with different physical conditions.
\end{itemize}

\section{Methodology}

\begin{figure}[t!]
\centering
\includegraphics[width=1\linewidth]{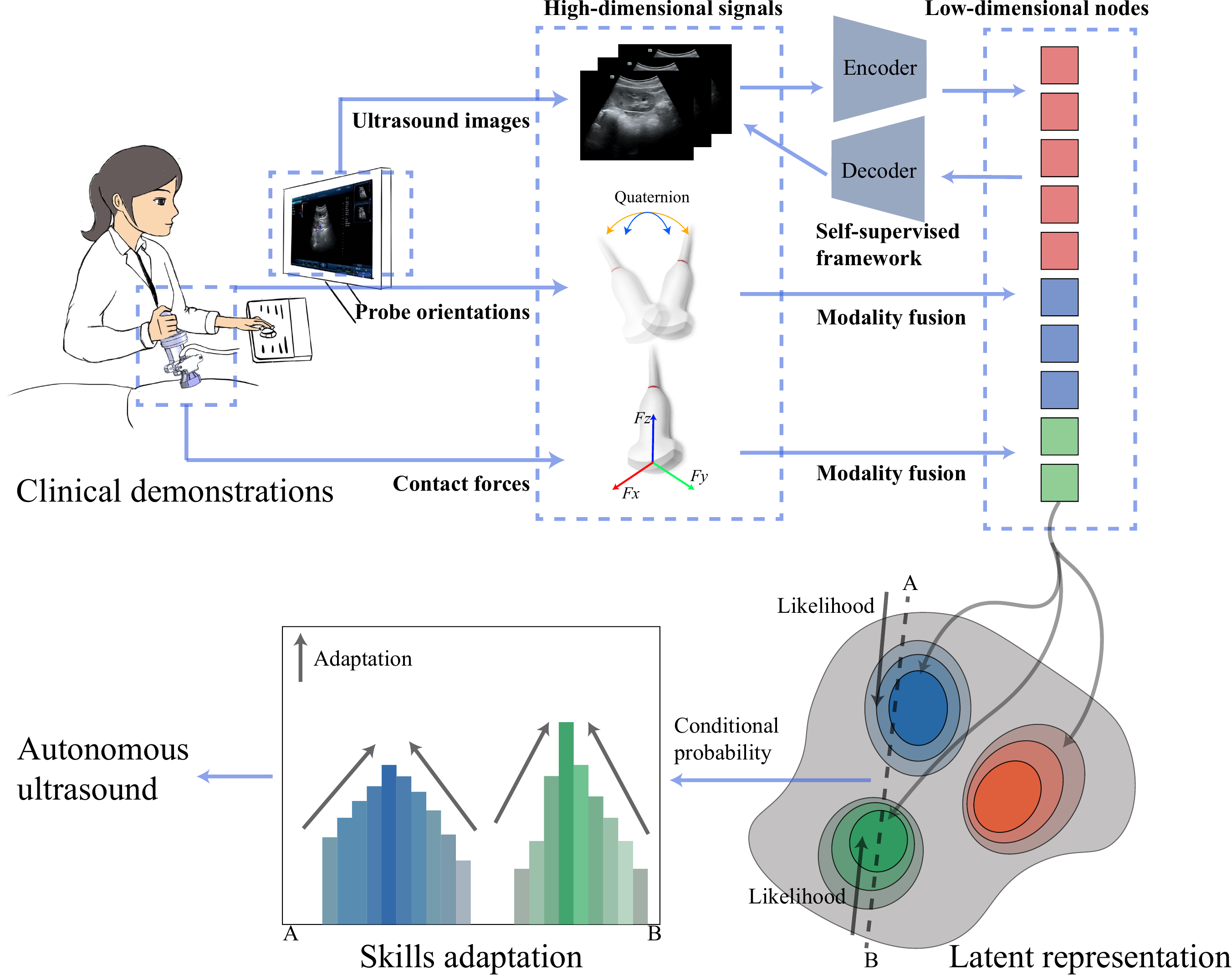}
\caption{\footnotesize The framework of latent task representation and robotic skills adaptation. Synchronous sequences are demonstrated and embedded into latent space as discrete nodes. The global distribution is learned by a probability model. The latent state's likelihood can be regarded as an evaluation. Any low-likelihood nodes will be judged as unstable, and an adaptive optimizer will fine-tune them to near and stable predictions in high-confidence regions.}
\vspace{-5mm}
\label{fig:framework}
\end{figure}

Fig.~\ref{fig:framework} presents the framework of methodology in this paper. Our goal is to learn freehand ultrasound skills from professional sonographers' demonstrations and further generalize them to some kind of adaptive robotic ultrasound skills for patients of different ages, genders, and body physiques.

\subsection{Problem Formulation}

The data involved in the training process is as follows:
\begin{itemize}
\item $D=\{d_1,d_2,...,d_N\}=\{(x_i,p_i,f_i)\}_{i=1...N}$ denotes a high-dimensional dataset with $N$ observations.
\item $x_i\in\mathbb{R}^{224\times224\times1}$ denotes the $i$-th collected ultrasound image with cropped size.
\item $p_i\in\mathbb{R}^4$ denotes the probe orientation (quaternion).
\item $f_i\in\mathbb{R}^6$ denotes the contact force/torque.
\end{itemize}
Additionally, $w_i=\{p_i,f_i\}$ is the variable directly related to robot control. To perform autonomous ultrasound based on a prior dataset $D$, the target function is described as:
\begin{equation}
    \begin{aligned}
        \max \mathbb{E}[&logP(x_{t},\hat{w}_{t+1}|D)] \\
        s.t. \quad &\hat{w}_{t+1} = f_{AU}(x_t,w_t|D),
    \end{aligned}
\label{eq:formulation}
\end{equation}
where $\hat{w}_{t+1}$ is the prediction, $f_{AU}(\cdot)$ denotes the robotic policy, and $P(\cdot)$ represents the conditional probability.

\subsection{Latent Representation and Skills Adaptation}

To latently represent the autonomous ultrasound task, we employ self-supervised learning to determine the mapping rule from high- to low-dimensional spaces, and $v_i$ denotes the feature vector of $x_i$. The represented demonstrations are equivalent to $D^{'}=\{(v_i,p_i,f_i)\}_{i=1...N}$. To this end, demonstrations are replaced by nodes (concatenations of $v$, $p$, and $f$) in latent space, and learning ultrasound skills is equal to finding a proper distribution to measure the conditional probability in Eq.~\ref{eq:formulation}. We assume nodes in latent space follow the Gaussian mixture distribution:
\begin{equation}
P(D^{'}|\Omega) = \sum_{k=1}^{K} \pi_k \mathcal{N}(D^{'}|\mu_k,\Sigma_k).
\label{eq:gmm}
\end{equation}
where $\pi_k$ is the prior of $k$-th Gaussian component, $\mathcal{N}(\mu_k,\Sigma_k)$ is the Gaussian distribution with mean $\mu_k$ and covariance $\Sigma_k$.

\begin{figure}[t!]
\centering
\includegraphics[width=1\linewidth]{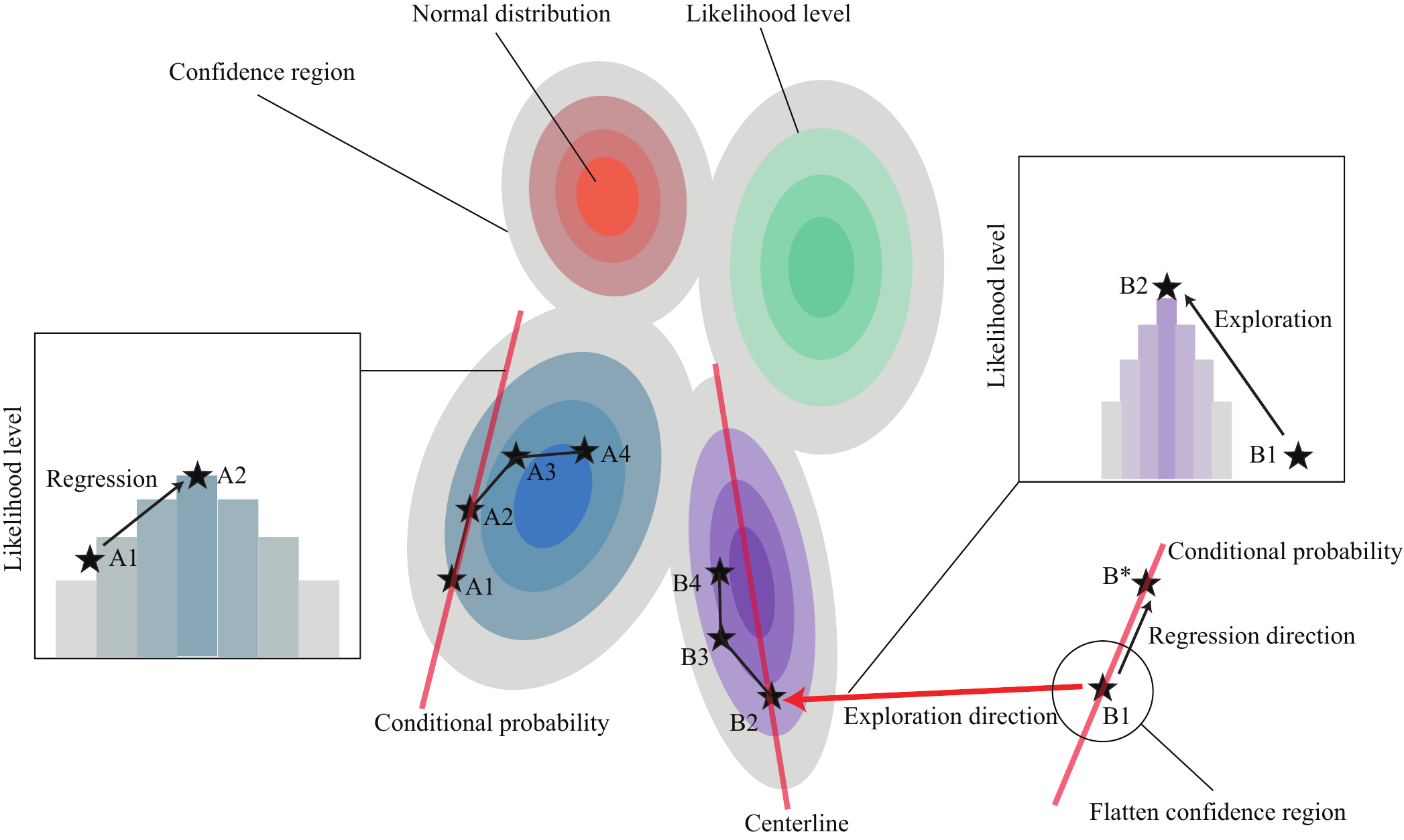}
\caption{\footnotesize The adaptation scheme of learned ultrasound skills in this paper. If an embedded state is mapped to a high-likelihood node, such as $A1$, then the state will be judged as stable. In this case, the optimization is a step-by-step iterative process. In another case, the embedded state is in a low-likelihood region and judged as unstable, such as $B1$. For this node, the local likelihood does not change that much and the confidence tends to be small and flat. The singularity will be straightly fixed to the nearest Gaussian distribution (from $B1$ to $B2$) and avoid following the gradient of conditional probability, which usually produces a relatively random result (from $B1$ to $B^{*}$).}
\vspace{-5mm}
\label{fig:adaptation}
\end{figure}

\subsubsection{Prediction Step}

The learned ultrasound skills can be reproduced by Gaussian Mixture Regression (GMR) \cite{billard2008robot}, and the conditional expectation of $w$ given $v$ is defined as:
\begin{equation}
\begin{aligned}
& P(w|v) \sim N(\hat{\mu}^w,\hat{\Sigma}^{w,w}), \\
& \hat{\mu}^w = \sum_{k=1}^K \pi_k \hat{\mu}_k^w, \quad \hat{\Sigma}^{w,w} = \sum_{k=1}^{K} \pi_k^2 \hat{\Sigma}_k^{w,w}.
\end{aligned}
\label{eq:gmr}
\end{equation}

\subsubsection{Evaluation Step}

The likelihood range $[a_k^{m\sigma},b_k^{m\sigma}]$ of $k$-th Gaussian component at $m$ standard deviation is:
\begin{equation}
\begin{aligned}
a_k^{m\sigma} &= \min Lik_k^{m\sigma}(d), \\
b_k^{m\sigma} &= \max Lik_k^{m\sigma}(d).
\end{aligned}
\label{eq::boundary}
\end{equation}
The stability of $\hat{d}=\{v,\hat{w}\}$ can be measured by:
\begin{equation}
\begin{aligned}
Case 1: Lik_k(\hat{d}) \in [a_k^{m\sigma},b_k^{m\sigma}], \exists{k} \in [1,2,...,K] \\
Case 2: Lik_k(\hat{d}) \notin [a_k^{m\sigma},b_k^{m\sigma}], \forall{k} \in [1,2,...,K]
\end{aligned}
\label{eq::stable}
\end{equation}
If the prediction's likelihood is subject to $Case1$, it will be considered stable. Else if its likelihood is subject to $Case2$, the prediction will be considered unstable and needs to be fine-tuned in the following adaptation stage.

\subsubsection{Adaptation Step}

The adaptation scheme is shown in Fig.~\ref{fig:adaptation}. For unstable cases, selecting the GMR result in Eq.~\ref{eq:gmr} as the local optimum may be invalid. That is because similar samples are scarce in demonstrations, and the probability model is non-referential to singularities. To avoid this, $\hat{w}$ will be adjusted through the local exploration \cite{li2014learning}:
\begin{equation}
    \hat{w} = \operatorname*{arg\,min}_{\mu_k^w} l_k \quad s.t. \quad k={1,2,...,K},
\label{eq:adaptation}
\end{equation}
where $l_k$ is the Mahalanobis distance between the prediction and the center of $k$-th Gaussian component.

\section{Experiments}

\subsection{Clinical Demonstrations}

\begin{figure}[t!]
\centering
\includegraphics[width=1\linewidth]{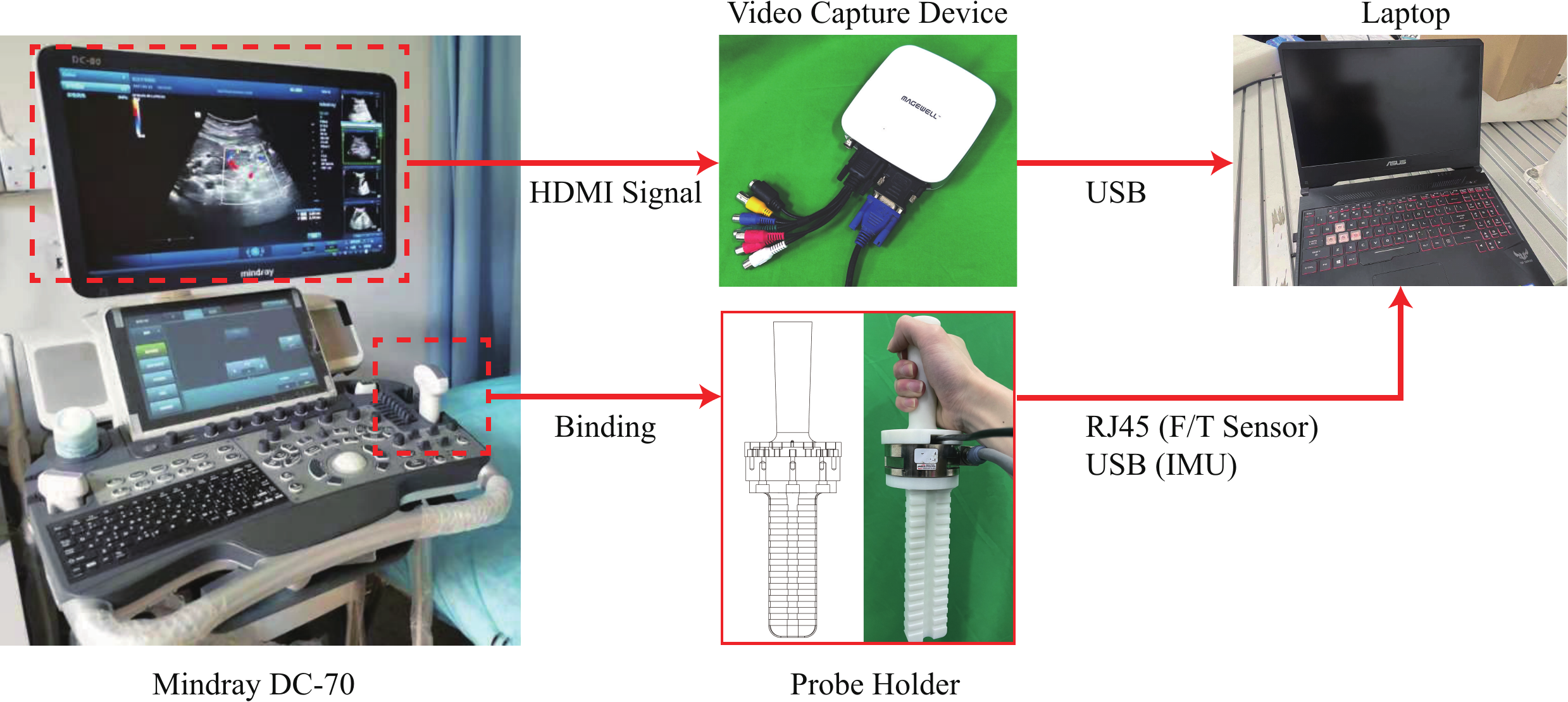}
\caption{\footnotesize The hardware system for clinical experiments.}
\vspace{-5mm}
\label{fig:hardware}
\end{figure}


This work was supported and supervised by the Medical Ethics Committee, School of Medicine, Wuhan University, Ethics Statement No.WHU2021-PMC002. Our experimental setup is shown in Fig.~\ref{fig:hardware}. The clinical experiment was at the Hospital of Wuhan University. The sonographer performed 5 times of left kidney examinations for each volunteer. The recording frequency was 10 Hz. A whole demonstration was started with the sonographer vertically holding the probe holder upon the target part. After $3\sim5$ seconds from the coordinate alignment and gravity calibration, the sonographer would start to perform freehand ultrasound examinations. Each demonstration lasted $40\sim80$ seconds. A total of 24 volunteers (14 males and 10 females) participated in this experiment. In order to make the individual differences in demonstrations more typical, we not only considered gender but also deliberately included people of different degrees of obesity and age groups in the experiment. Volunteers' BMI ranged from 16.4 (underweight) to 26.7 (overweight), and their ages ranged from 19 to 67. Totally, there were 53571 sets of multimodal data in our dataset.

\subsection{Latent Representation}

\begin{figure}[t!]
\centering
\includegraphics[width=1\linewidth]{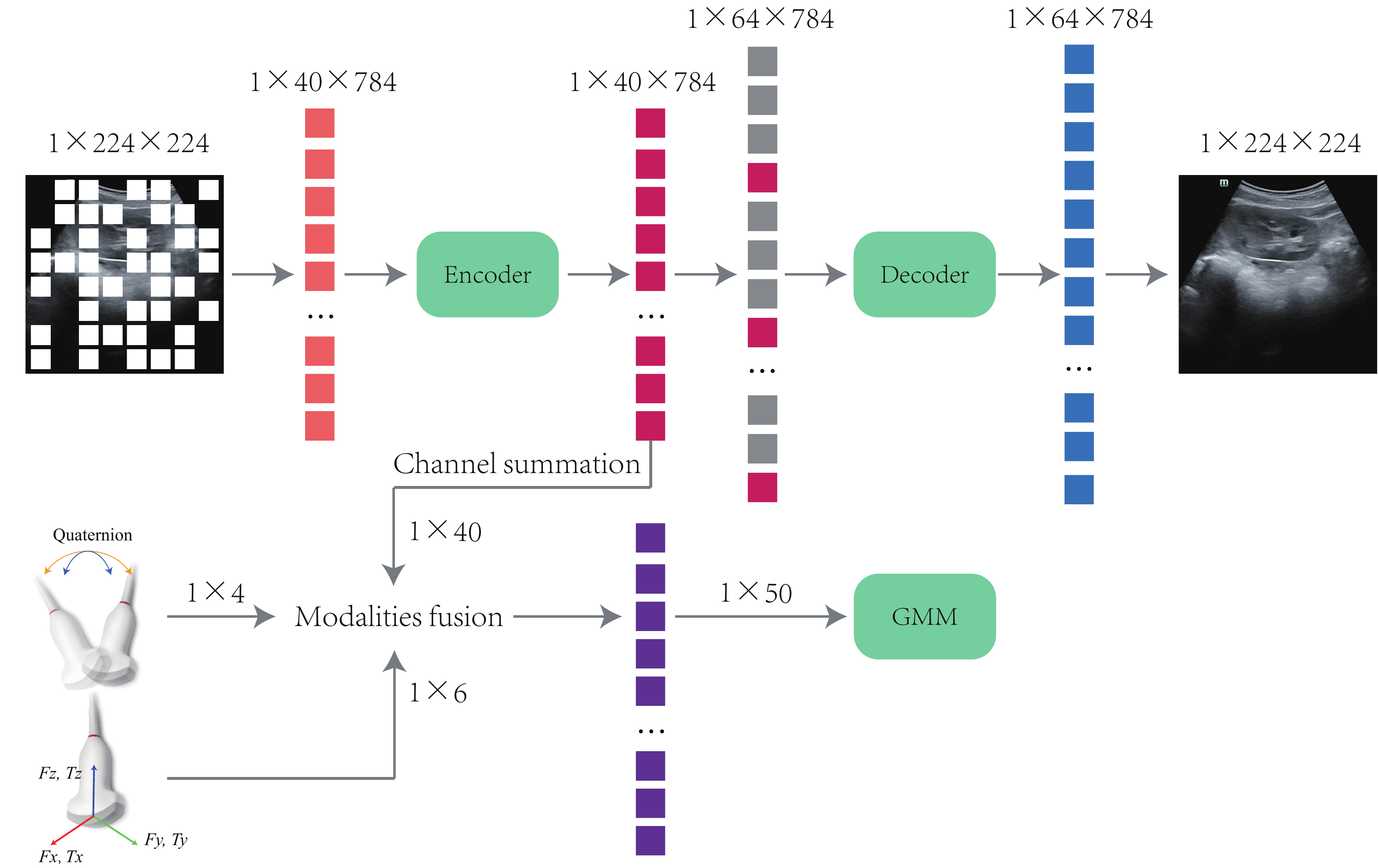}
\caption{\footnotesize The modalities fusion framework. The encoder-decoder backbone is performed by Masked Autoencoder \cite{he2022masked}, and fusion is performed by channel summation and concatenating.}
\vspace{-5mm}
\label{fig:fusion}
\end{figure}

The ultrasound skills model first learns to encode ultrasound images into feature vectors and then concatenates three modalities as low-dimensional nodes. Then, the embedded nodes in latent space are encapsulated into a Gaussian Mixture Model (GMM). The modalities fusion pipeline is presented in Fig.~\ref{fig:fusion}. In detail, the ultrasound images' feature extraction is performed by Masked Auto Encoder (MAE) \cite{he2022masked}. The raw ultrasound images are with a $1920\times1080$ resolution. Through cropping, scaling, and grayscale, they are converted to $1\times224\times224$ preprocessed images. Each preprocessed one is divided into $8\times8$ patches with the size of $28\times28$. A random selector picks 40 out of 64 patches and the rest are masked out. The remaining 40 patches are reshaped to a $1\times40\times784$ feature vector, and thus processed by the encoder and decoder modules.

The $1\times40\times784$ feature vector is too large to merge with the other modalities. Therefore, a channel summation is added to compress 784 values in each channel into one single feature value, and the ultrasound modality is replaced by a $1\times40$ feature vector. Now, by concatenating the $1\times40$ feature vector, $1\times4$ quaternion vector, and $1\times6$ force/torque vector, every state in original demonstrations can be mapped to a node in 50D latent space. To model these low-dimensional nodes as a probability distribution, we use the GMM with 16 Gaussian components (about 40000 parameters).

\subsection{Skills Adaptation}

\begin{table}[b]
\centering
\scriptsize
\vspace{-5mm}
\caption{\footnotesize Summary results of MC and GMM methods in five tasks.}
\label{tab:exp_results}
\renewcommand\arraystretch{1.5}
\setlength{\tabcolsep}{1mm}{
\begin{tabular}{cccccc}
\toprule
\multicolumn{2}{c}{Methods} & \makecell[c]{Pose error\\(degree)} & \makecell[c]{Force error\\(N)} & \makecell[c]{Torque error\\(Nm)} & FPS \\
\midrule
\multirow{8}{*}{MC}
& 50 samples    & $27.36\pm26.08$ & $2.05\pm0.86$ & $0.14\pm0.06$ & 66.07 \\
& 100 samples   & $26.43\pm25.90$ & $2.04\pm0.85$ & $0.14\pm0.06$ & 43.75 \\
& 200 samples   & $26.65\pm26.07$ & $1.94\pm0.83$ & $0.14\pm0.06$ & 34.97 \\
& 500 samples   & $25.55\pm25.43$ & $1.98\pm0.85$ & $0.14\pm0.06$ & 15.00 \\
& 1000 samples  & $24.69\pm25.32$ & $1.92\pm0.85$ & $0.14\pm0.06$ & 7.40 \\
& 2000 samples  & $27.06\pm26.26$ & $2.01\pm0.87$ & $0.13\pm0.06$ & 3.64 \\
& 5000 samples  & $23.63\pm25.17$ & $1.87\pm0.83$ & $0.13\pm0.06$ & 1.36 \\
& 10000 samples & $25.24\pm25.79$ & $2.29\pm0.91$ & $0.13\pm0.06$ & 0.61 \\ \cline{2-6} 
\multirow{3}{*}{\makecell[c]{GMM\\(Ours)}}
& 1-sigma region & $20.63\pm12.29$ & $1.22\pm0.47$ & $0.10\pm0.05$ & 304.98 \\
& 2-sigma region & $20.64\pm12.17$ & $1.21\pm0.45$ & $0.09\pm0.05$ & 312.01 \\
& 3-sigma region & $20.61\pm12.11$ & $1.21\pm0.46$ & $0.09\pm0.04$ & 306.22 \\
\bottomrule
\end{tabular}}
\end{table}

\begin{figure}[t!]
\centering
\subfloat[\footnotesize Pose error]{\includegraphics[width=1\linewidth, trim=0 0 0 25, clip]{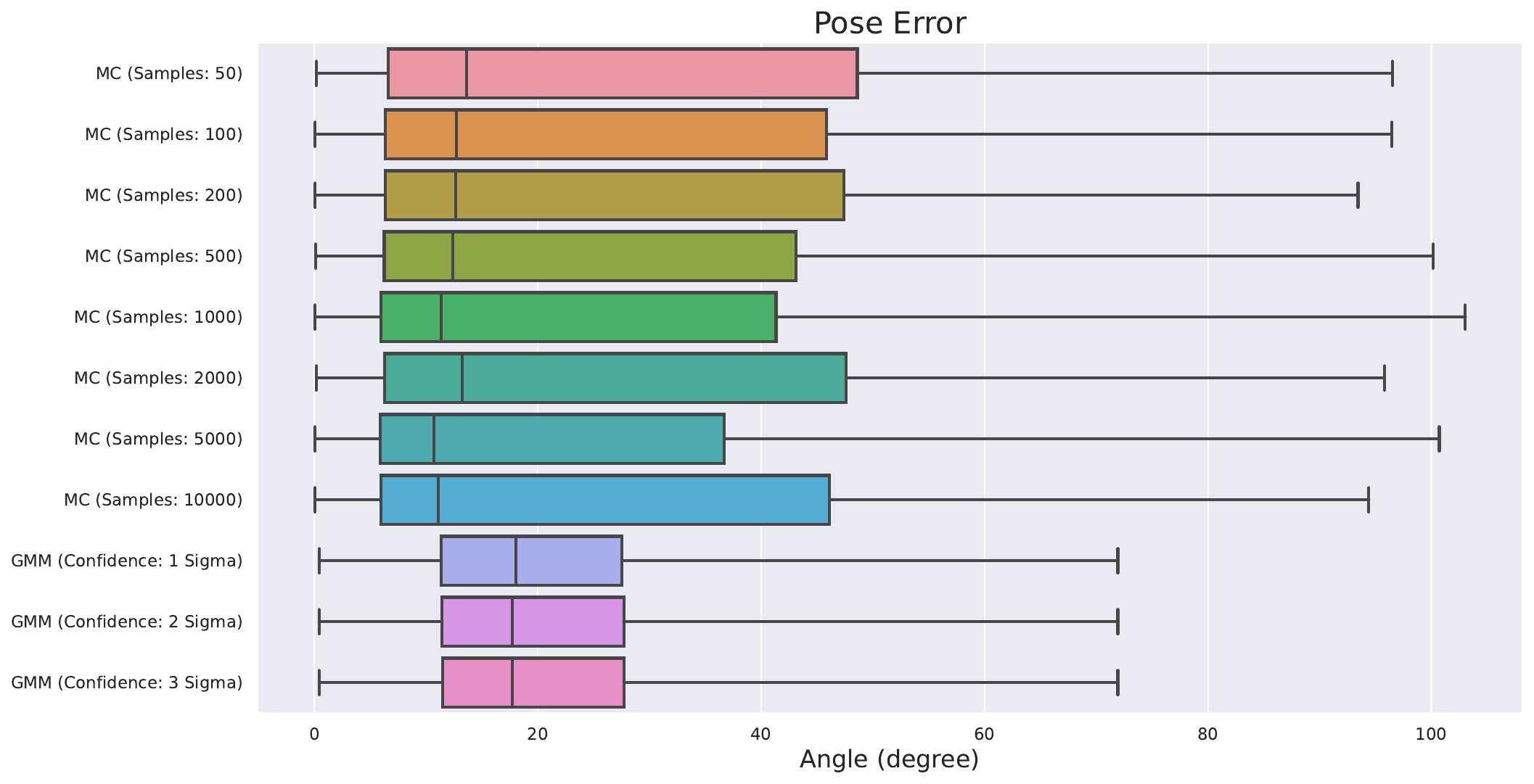}}
\quad
\subfloat[Force error]{\includegraphics[width=1\linewidth, trim=0 0 0 25, clip]{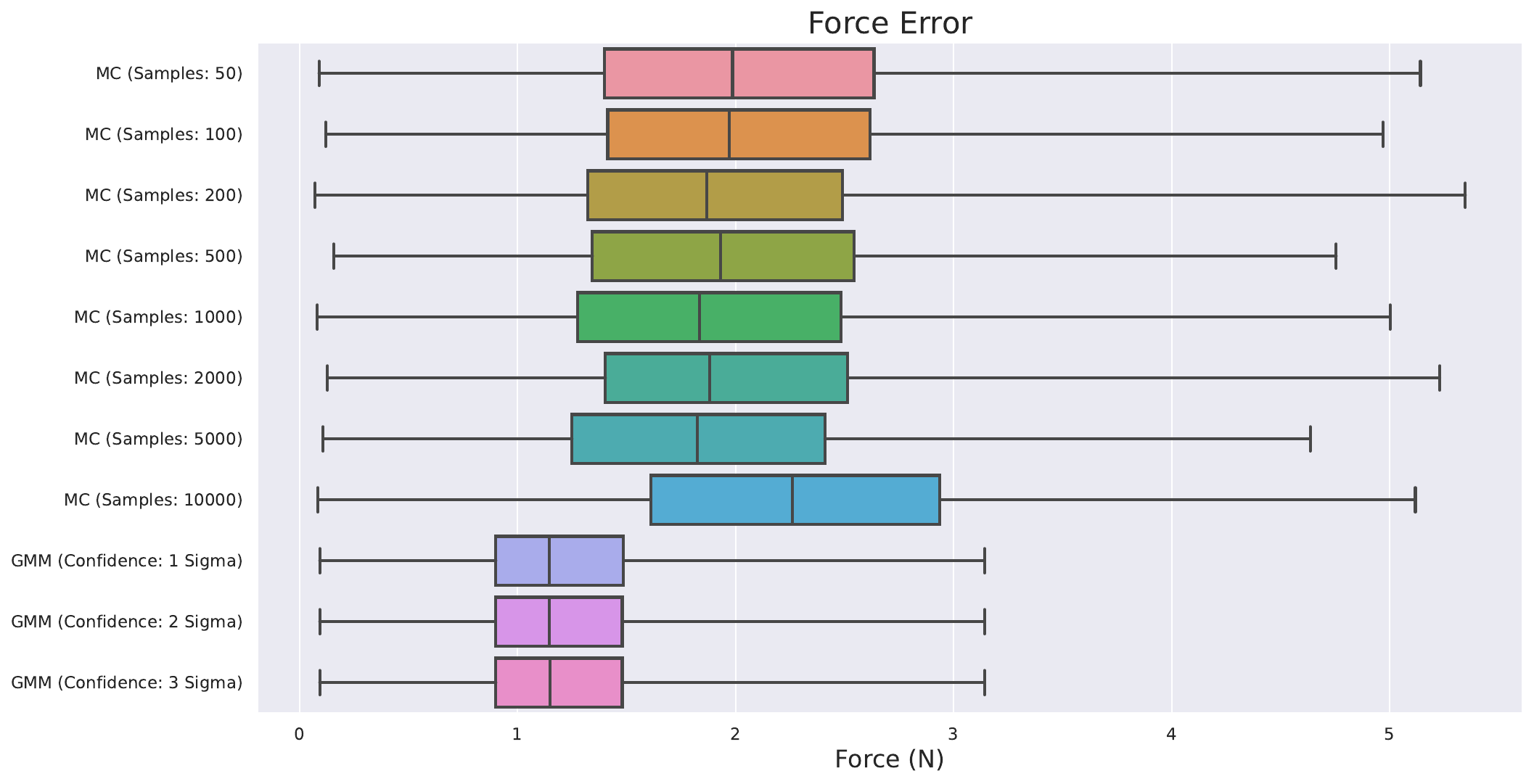}}
\quad
\subfloat[Torque error]{\includegraphics[width=1\linewidth, trim=0 0 0 25, clip]{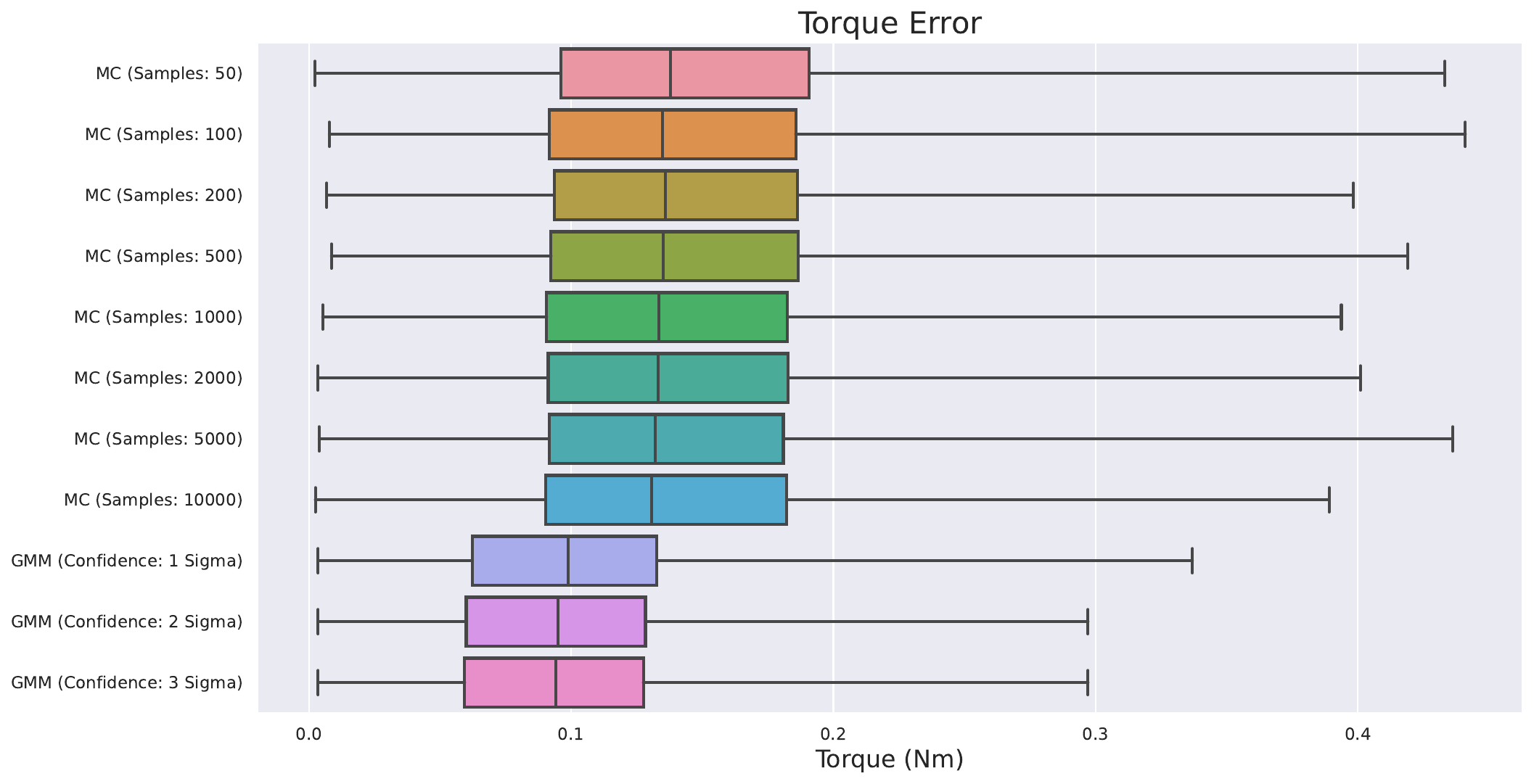}}
\caption{\footnotesize Boxplots of MC and GMM methods in five tasks.}
\vspace{-5mm}
\label{fig:exp_results}
\end{figure}

\begin{figure}[t!]
\centering
\subfloat[No.1 volunteer, male, age 19, BMI 17.99 (underweight).]{\includegraphics[width=1\linewidth]{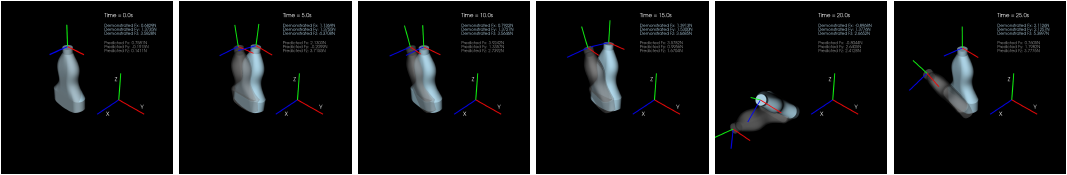}}
\quad
\subfloat[No.5 volunteer, male, age 23, BMI 26.58 (overweight).]{\includegraphics[width=1\linewidth]{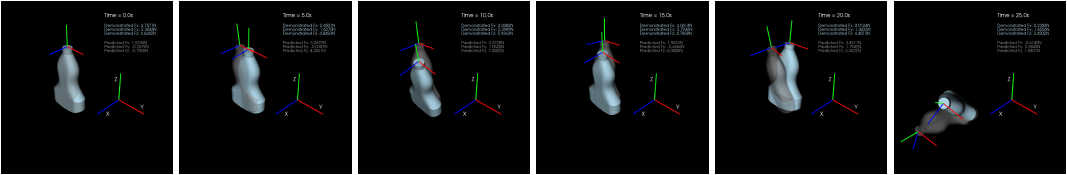}}
\quad
\subfloat[No.12 volunteer, male, age 36, BMI 19.13 (normal).]{\includegraphics[width=1\linewidth]{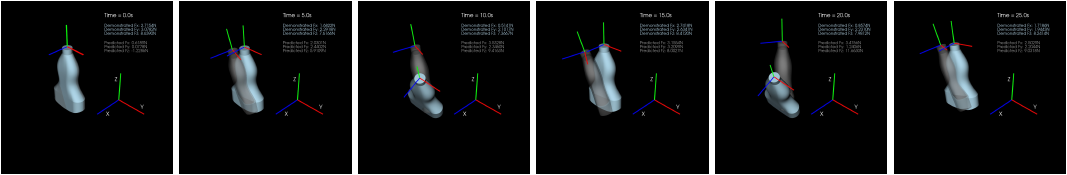}}
\quad
\subfloat[No.13 volunteer, female, age 67, BMI 25.39 (overweight).]{\includegraphics[width=1\linewidth]{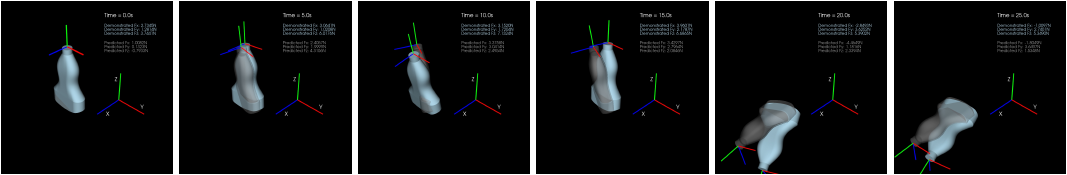}}
\quad
\subfloat[No.15 volunteer, female, age 22, BMI 22.86 (normal).]{\includegraphics[width=1\linewidth]{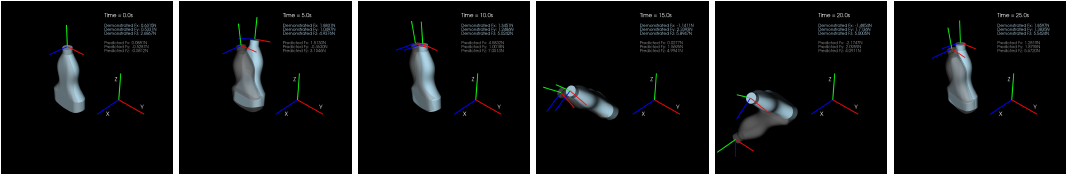}}
\caption{\footnotesize Visualization of autonomous ultrasound tasks.}
\vspace{-5mm}
\label{fig:visualization}
\end{figure}

Once parameters of MAE and GMM are learned, the probability model in Eq.~\ref{eq:formulation} is equally determined. In this way, autonomous ultrasound can be reproduced by GMR. However, singularities are inevitable since demonstrations are always insufficient to fully cover all cases. To this end, we propose the robotic skills adaptation to alleviate this problem, which computes the likelihood bounds for every Gaussian distribution in $m$ standard deviation, and fine-tunes those unstable nodes with low likelihoods.

Our baseline is the Monte Carlo (MC) method \cite{deng2021scis, li2022learning}, where the probability model is replaced by a Multilayer Perceptron (MLP) model. Experimental results have shown that the MC method could learn demonstrated ultrasound skills and reproduce an expert-like strategy. But it tends to sacrifice more computing resources to acquire more acceptable results. Therefore, we study the upper limitation of the MC method and make a quantitative comparison with the GMM method in this experiment. In detail, $1\sigma$, $2\sigma$, and $3\sigma$ confidence regions are chosen for the GMM method respectively. While the numbers of samples are 50, 100, 200, 500, 1000, 2000, 5000, and 10000 in the MC method. For the intra-patient experiment, the first 4 demonstrations of every volunteer are used for training and the last for evaluation. For the inter-patient experiment, all demonstrations of $1\sim22$ volunteers are used for training, and those of $23\sim24$ volunteers are used for evaluation. In addition, we further evaluate the proposed method in different tasks (inter-gender, inter-age, and inter-BMI tasks) with significantly larger individual differences to explore our method's robustness and generalization. Summary results are presented in Fig.~\ref{fig:exp_results} and TABLE.~\ref{tab:exp_results}. The $3\sigma$ GMM policy (gray) is visualized in Fig.~\ref{fig:visualization}, which shows great agreement with the sonographer's demonstrations (blue).

\section{Conclusion}

Learning autonomous strategies for robotic ultrasound systems is still unsolved, which is mainly related to the absence of two fundamental capabilities: a proper representation method to model and reproduce, and an adaptive method to generalize and adjust the learned ultrasound skills. In this paper, we propose to learn robotic ultrasound skills from clinical freehand examinations demonstrated by a professional sonographer with latent representation and skills adaptation, which highlights feasibility and generalization for patients with different physical conditions. During the offline stage, the multimodal ultrasound skills are encapsulated into a low-dimensional probability model through a fully self-supervised framework, which takes ultrasound images, probe orientations, and contact forces into account. During the online stage, the optimal prediction will be selected and evaluated by the probability model. For any singularities with unstable and uncertain states, the adaptive optimizer fine-tunes them to near and stable predictions in high-confidence regions. The experimental results in intra-patient, inter-patient, inter-gender, inter-age, and inter-BMI tasks show that the proposed approach can generate complex strategies for diverse populations, with accuracy and speed significantly better than our previous method.

\section*{Acknowledgments}


This work was supported by the Fundamental Research Funds for the Central Universities (No.2042023KF0110).

\newpage

\bibliographystyle{plainnat}
\bibliography{references}

\end{document}


\title{Supplementary Materials: Learning Autonomous Ultrasound via Latent Task Representation and Robotic Skills Adaptation}

\author{\authorblockN{Xutian Deng\authorrefmark{1},
Junnan Jiang\authorrefmark{2},
Wen Cheng\authorrefmark{3} and
Miao Li\authorrefmark{4}}
\authorblockA{\authorrefmark{1}School of Computer Science, Wuhan University, Wuhan 430072, China\\
Email: \href{mailto:dengxutian@whu.edu.cn}{dengxutian@whu.edu.cn}}
\authorblockA{\authorrefmark{2}School of Power and Mechanical Engineering, Wuhan University, Wuhan 430072, China}
\authorblockA{\authorrefmark{3}Hospital of Wuhan University, Wuhan University, Wuhan 430072, China}
\authorblockA{\authorrefmark{4}School of Microelectronics, Wuhan University, Wuhan 430072, China\\
Email: \href{mailto:miao.li@whu.edu.cn}{miao.li@whu.edu.cn}}
}

\maketitle

\section*{Supplementary \#1: Physical information of volunteers}

\begin{table}[htbp]
\centering
\caption{Physical condition of 24 clinical volunteers.}
\renewcommand\arraystretch{1.5}
\setlength{\tabcolsep}{6mm}{
\begin{tabular}{ccccccc}
\toprule
Volunteer & Age & Gender & Height (m) & Weight (kg) & BMI & Assessment  \\
\midrule
1  & 19  & Male   & 1.65 & 49 & 17.9982 & Underweight \\
2  & 35  & Male   & 1.74 & 68 & 22.4600 & Normal      \\
3  & 27  & Male   & 1.72 & 79 & 26.7036 & Overweight  \\
4  & 25  & Male   & 1.72 & 62 & 20.9573 & Normal      \\
5  & 23  & Male   & 1.84 & 90 & 26.5832 & Overweight  \\
6  & 24  & Male   & 1.62 & 46 & 17.5278 & Underweight \\
7  & 23  & Male   & 1.79 & 81 & 25.2801 & Overweight  \\
8  & 23  & Male   & 1.76 & 53 & 17.1100 & Underweight \\
9  & 22  & Male   & 1.77 & 80 & 25.5354 & Overweight  \\
10 & 24  & Male   & 1.82 & 72 & 21.7365 & Normal      \\
11 & 27  & Male   & 1.81 & 68 & 20.7564 & Normal      \\
12 & 36  & Male   & 1.68 & 54 & 19.1327 & Normal      \\
13 & 67  & Female & 1.60 & 65 & 25.3906 & Overweight  \\
14 & 24  & Male   & 1.70 & 67 & 23.1834 & Normal      \\
15 & 22  & Female & 1.62 & 60 & 22.8624 & Normal      \\
16 & 23  & Male   & 1.70 & 55 & 19.0311 & Normal      \\
17 & 19  & Female & 1.58 & 46 & 18.4265 & Underweight \\
18 & 24  & Female & 1.63 & 51 & 19.1953 & Normal      \\
19 & 25  & Female & 1.55 & 55 & 22.8928 & Normal      \\
20 & 21  & Female & 1.69 & 60 & 21.0077 & Normal      \\
21 & 19  & Female & 1.54 & 39 & 16.4446 & Underweight \\
22 & 19  & Female & 1.61 & 55 & 21.2183 & Normal      \\
23 & 24  & Female & 1.58 & 50 & 20.0288 & Normal      \\
24 & 25  & Female & 1.62 & 49 & 18.6709 & Normal      \\
\bottomrule
\end{tabular}}
\end{table}

\newpage
\section*{Supplementary \#2: Hardware Details}

\begin{figure}[htbp]
\centering
\subfloat[CAD figure.]{\includegraphics[width=0.3\linewidth]{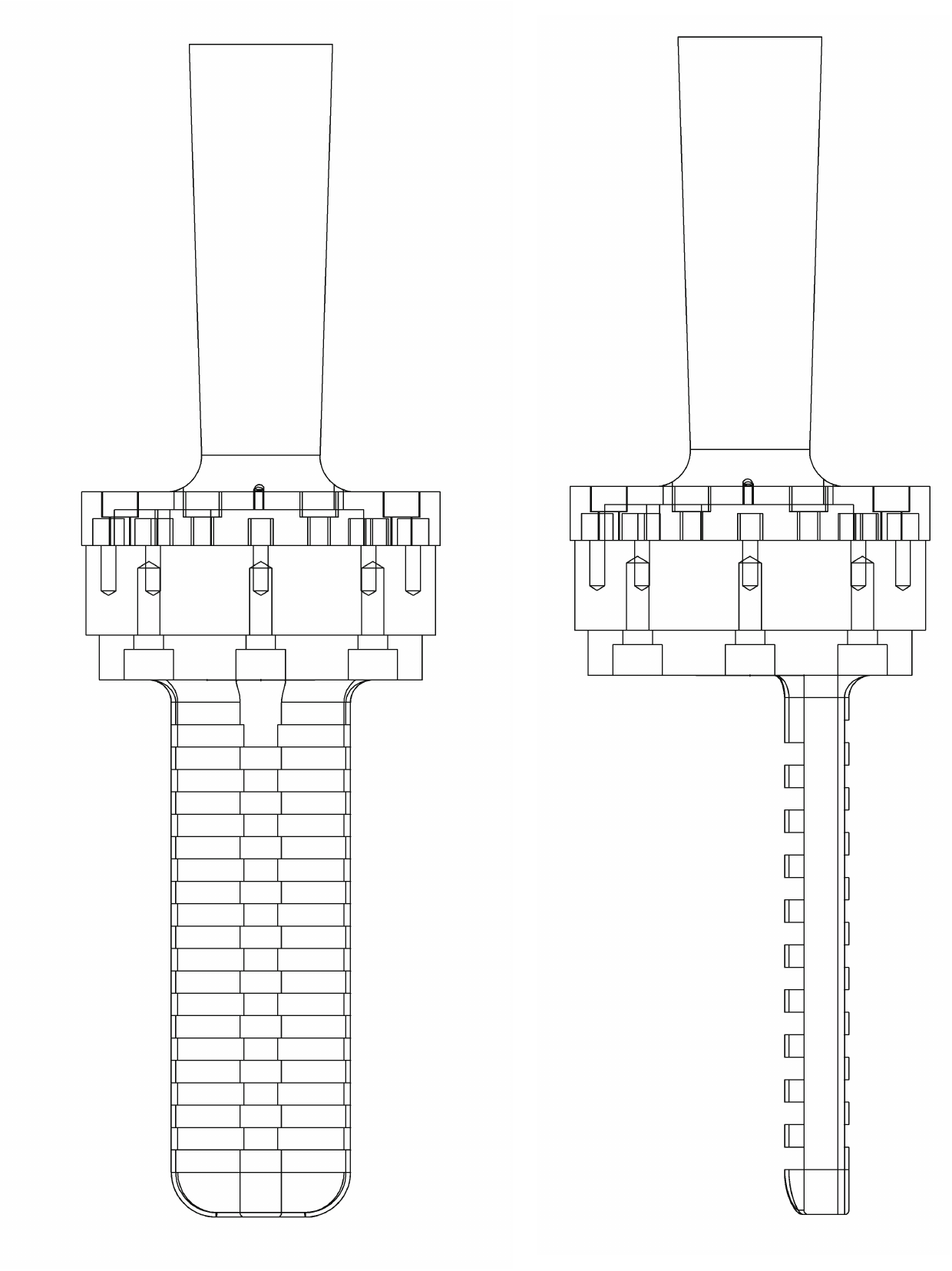}}
\subfloat[Assembled probe holder.]{\includegraphics[width=0.68\linewidth]{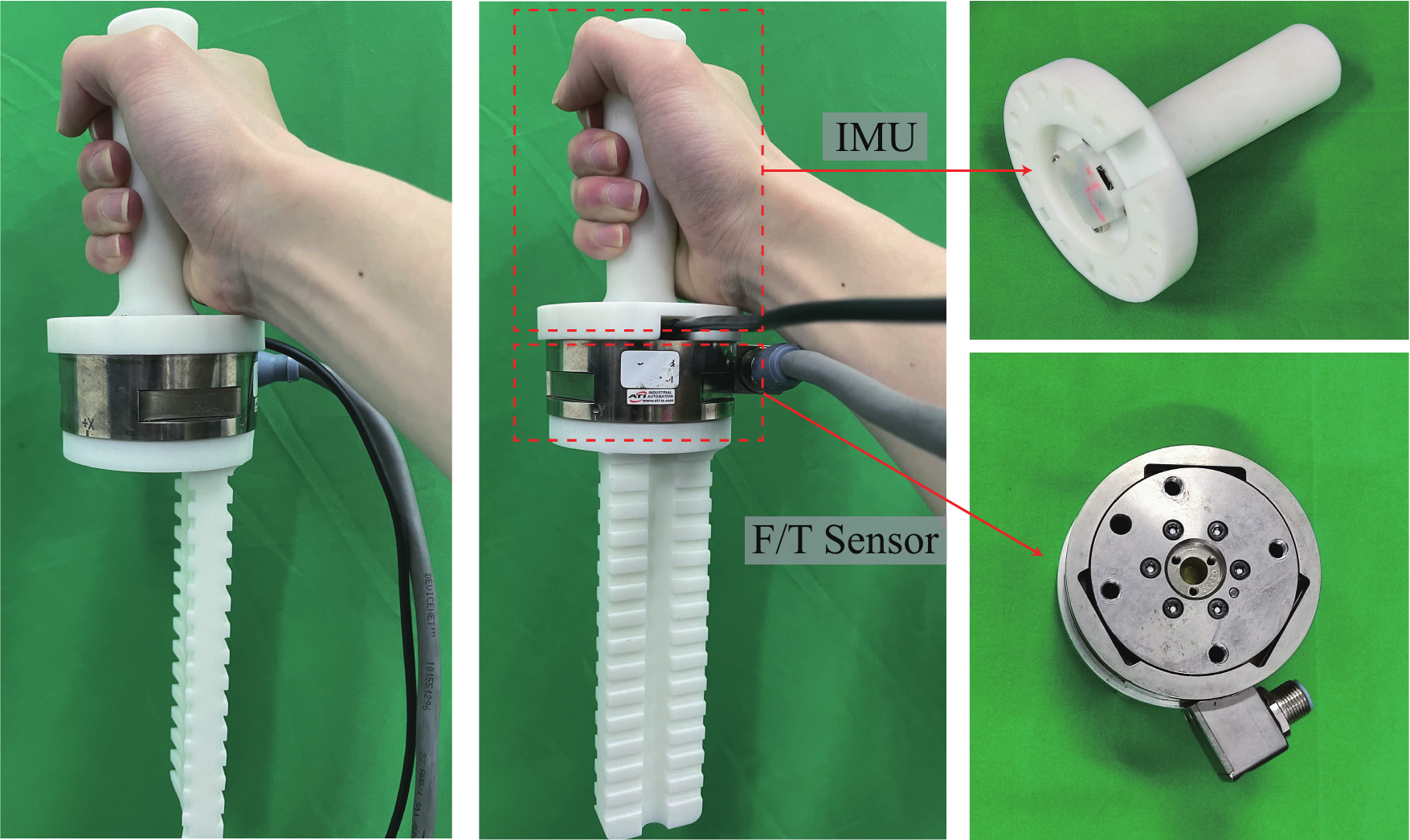}}
\caption{Some details about the probe holder in our experiment. This probe holder has two main sensors to monitor the real-time ultrasound demonstrations: a build-in IMU (ICM20948 module and up to 200 Hz) and a 6D force/torque sensor (ATI Gamma F/T sensor whose resolutions are 1/80 N, 1/80 N, and 1/40 N in three orthogonal directions). During the clinical experiment, the ultrasound probe will be mounted at the bottom, and the sonographer can hold the top handle to perform ultrasound examinations.}
\label{fig::hardware1}
\end{figure}

\newpage
\section*{Supplementary \#3: Experimental Snapshots}

\begin{figure}[htbp]
\centering
\subfloat[Stick figure.]{\includegraphics[width=0.48\linewidth]{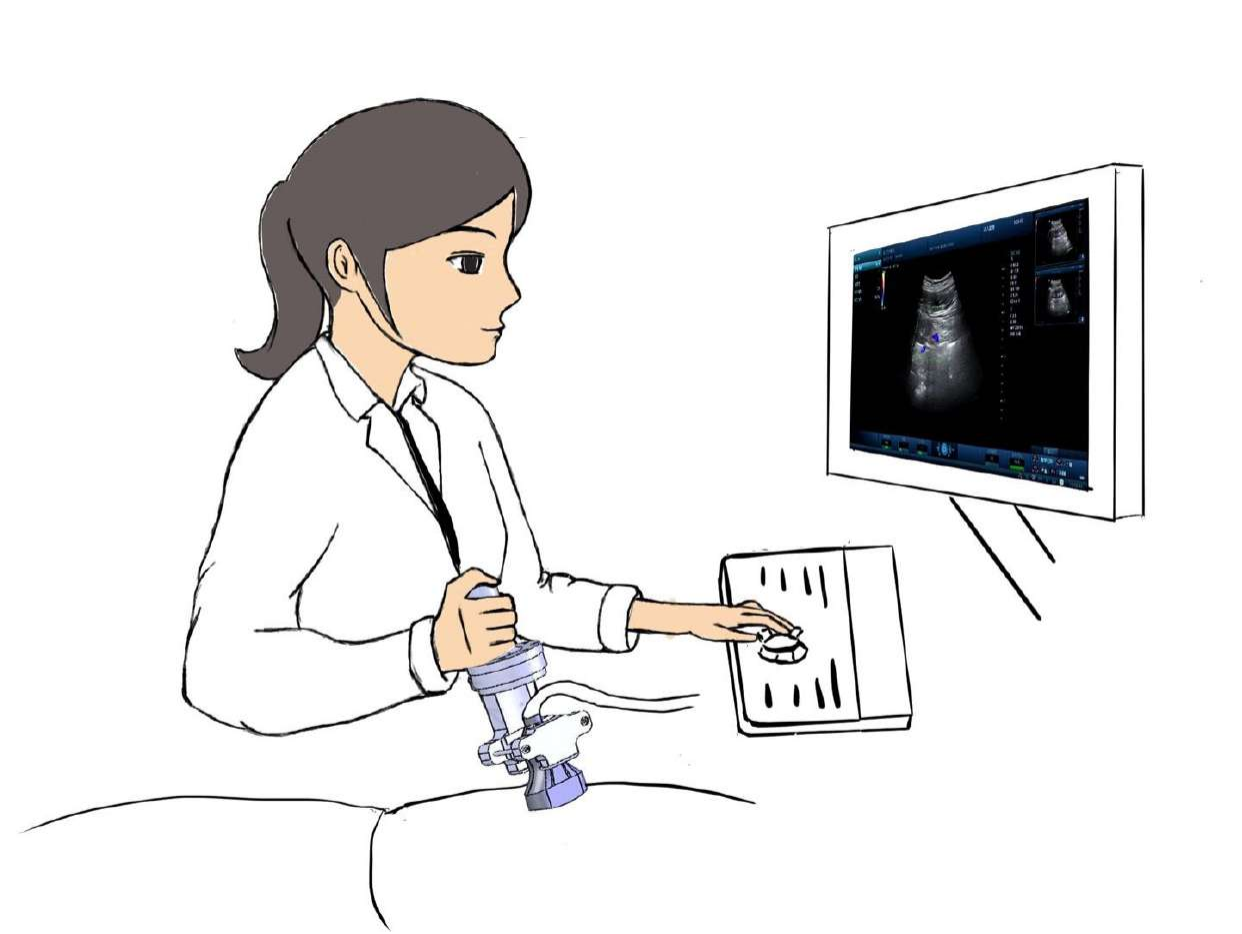}}
\subfloat[Clinical snapshot.]{\includegraphics[width=0.48\linewidth]{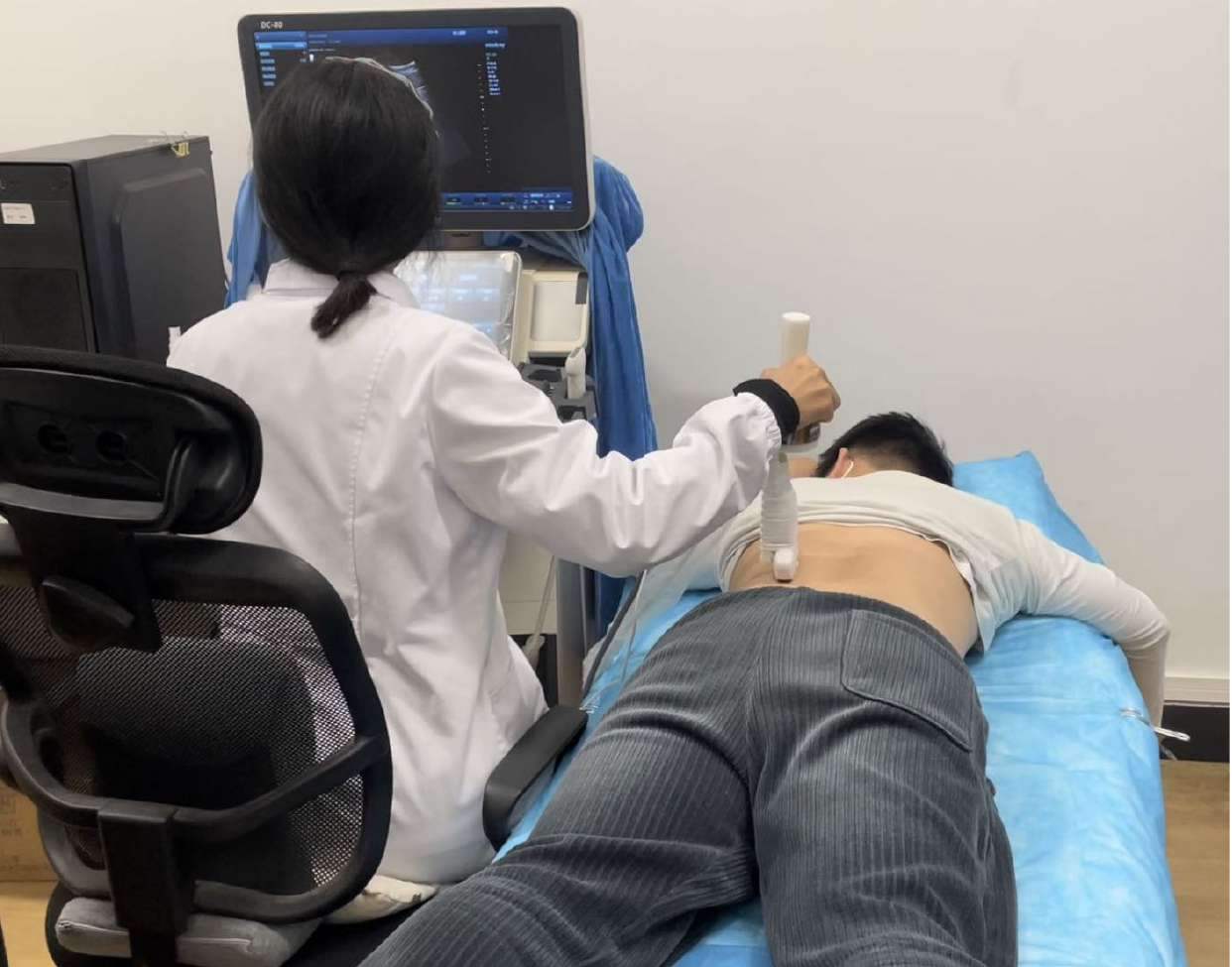}}
\quad
\subfloat[Freehand ultrasound tricks and the corresponding ultrasound images.]{\includegraphics[width=1\linewidth]{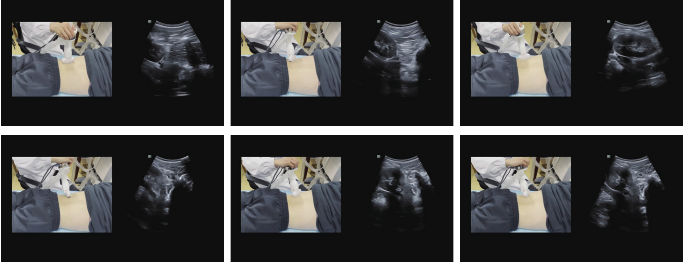}}
\caption{Some snapshots in clinical demonstration experiments. The instruction for hardware usage is shown in (a), by which the ultrasound videos, probe orientations, and contact forces were collected. An experimental snapshot is presented in (b). Some freehand ultrasound tricks of the sonographer and the corresponding ultrasound images are shown in (c). The whole demonstration started with the sonographer vertically holding the probe holder upon the target region. After $3\sim5$ seconds from the coordinate alignment and gravity calibration, the sonographer started to perform freehand ultrasound examinations. Each demonstration lasted $40\sim80$ seconds.}
\label{fig::snapshot}

\end{figure}

\newpage
\section*{Supplementary \#4: Visualized GMM}

\begin{figure}[htbp]
\centering
\includegraphics[width=1\linewidth]{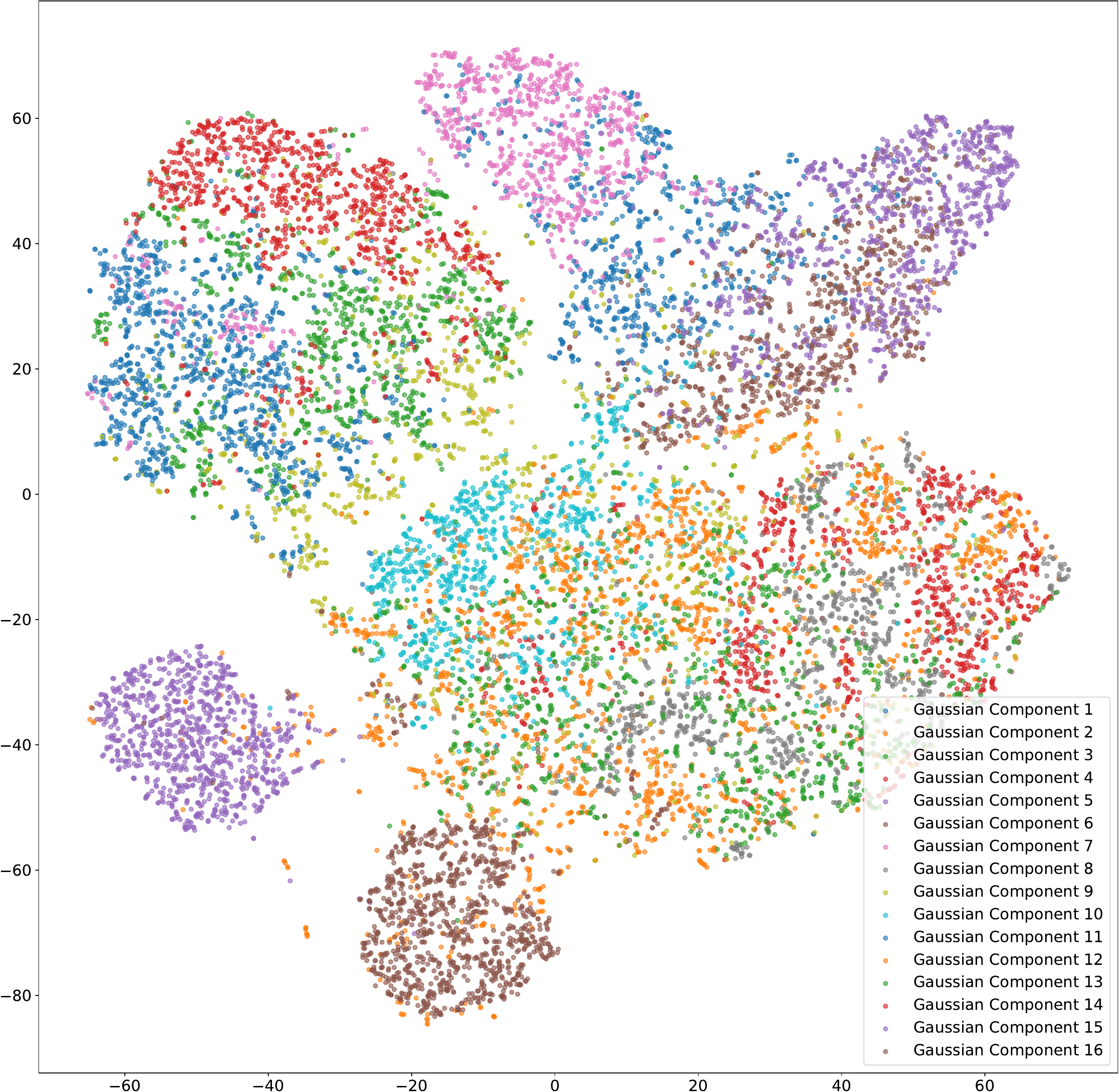}
\caption{Gaussian distributions visualized by the t-SNE algorithm in the 2D plane.}
\label{fig::scatter}
\end{figure}

\newpage
\section*{Supplementary \#5: Experimental Results of Inter-gender, Inter-age, and Inter-BMI Tasks}

\begin{table*}[htbp]
\centering
\renewcommand\arraystretch{1.5}
\setlength{\tabcolsep}{5mm}{
\caption{Experimental result of the inter-gender task.}
\begin{tabular}{cccccc}
\toprule
\multicolumn{2}{c}{Methods} & Pose error (degree) & Force error (N) & Torque error (Nm) & FPS \\
\midrule
\multirow{8}{*}{MC methods}
& 50 samples    & $29.3762\pm24.6150$ & $3.0254\pm1.2453$ & $0.2336\pm0.1082$ & 66.8829 \\
& 100 samples   & $29.0951\pm24.6043$ & $3.0084\pm1.2499$ & $0.2334\pm0.1089$ & 42.6738 \\
& 200 samples   & $29.0945\pm24.5366$ & $2.8858\pm1.2104$ & $0.2301\pm0.1075$ & 34.4192 \\
& 500 samples   & $29.0272\pm24.4550$ & $2.8912\pm1.1853$ & $0.2257\pm0.1058$ & 15.2769 \\
& 1000 samples  & $29.0807\pm24.5056$ & $2.9387\pm1.2339$ & $0.2256\pm0.1048$ & 7.3601 \\
& 2000 samples  & $28.4729\pm24.3151$ & $2.9162\pm1.2021$ & $0.2218\pm0.1049$ & 3.6720 \\
& 5000 samples  & $28.7922\pm24.4710$ & $2.9406\pm1.2192$ & $0.2202\pm0.1032$ & 1.3554 \\
& 10000 samples & $28.4612\pm24.2875$ & $2.8332\pm1.1765$ & $0.2184\pm0.1021$ & 0.6216 \\ \cline{2-6} 
\multirow{3}{*}{\makecell[c]{GMM methods\\(Ours)}}
& 1-sigma region & $18.8785\pm16.4591$ & $1.3750\pm0.6235$ & $0.1725\pm0.0783$ & 264.3962 \\
& 2-sigma region & $18.8595\pm16.1817$ & $1.3757\pm0.6182$ & $0.1725\pm0.0778$ & 271.3125 \\
& 3-sigma region & $18.8063\pm15.9940$ & $1.3750\pm0.6173$ & $0.1725\pm0.0775$ & 269.7957 \\
\bottomrule
\end{tabular}
\vspace{5mm}
\caption{Experimental result of the inter-age task.}
\begin{tabular}{cccccc}
\toprule
\multicolumn{2}{c}{Methods} & Pose error (degree) & Force error (N) & Torque error (Nm) & FPS \\
\midrule
\multirow{8}{*}{MC methods}
& 50 samples    & $27.0115\pm23.2189$ & $2.6255\pm1.2156$ & $0.1848\pm0.0874$ & 65.4180 \\
& 100 samples   & $26.4910\pm22.9918$ & $2.5866\pm1.1962$ & $0.1843\pm0.0872$ & 41.4879 \\
& 200 samples   & $26.5139\pm23.0421$ & $2.5071\pm1.1539$ & $0.1822\pm0.0866$ & 33.4342 \\
& 500 samples   & $26.0613\pm22.8256$ & $2.4649\pm1.1566$ & $0.1821\pm0.0852$ & 15.0523 \\
& 1000 samples  & $25.5738\pm22.7659$ & $2.4177\pm1.1043$ & $0.1799\pm0.0845$ & 7.6250 \\
& 2000 samples  & $25.6432\pm22.7773$ & $2.4564\pm1.1357$ & $0.1792\pm0.0838$ & 3.6656 \\
& 5000 samples  & $25.2520\pm22.6803$ & $2.5137\pm1.0770$ & $0.1782\pm0.0829$ & 1.3822 \\
& 10000 samples & $25.2013\pm22.4086$ & $2.4153\pm1.1250$ & $0.1774\pm0.0830$ & 0.6287 \\ \cline{2-6} 
\multirow{3}{*}{\makecell[c]{GMM methods\\(Ours)}}
& 1-sigma region & $15.5250\pm14.3686$ & $1.3589\pm0.8674$ & $0.1032\pm0.0567$ & 271.0660 \\
& 2-sigma region & $15.3933\pm14.3435$ & $1.3587\pm0.8602$ & $0.1020\pm0.0557$ & 272.9313 \\
& 3-sigma region & $15.3966\pm14.3325$ & $1.3577\pm0.8577$ & $0.1016\pm0.0555$ & 275.6897 \\
\bottomrule
\end{tabular}
\vspace{5mm}
\caption{Experimental result of the inter-BMI task.}
\begin{tabular}{cccccc}
\toprule
\multicolumn{2}{c}{Methods} & Pose error (degree) & Force error (N) & Torque error (Nm) & FPS \\
\midrule
\multirow{8}{*}{MC methods}
& 50 samples    & $29.8198\pm22.5163$ & $3.7173\pm1.6407$ & $0.2190\pm0.1035$ & 65.8205 \\
& 100 samples   & $29.8195\pm22.4535$ & $3.7914\pm1.6415$ & $0.2166\pm0.1030$ & 43.2587 \\
& 200 samples   & $29.6394\pm22.6435$ & $3.8630\pm1.6395$ & $0.2171\pm0.1013$ & 33.8700 \\
& 500 samples   & $29.2037\pm22.4719$ & $3.9510\pm1.6137$ & $0.2172\pm0.1025$ & 15.1218 \\
& 1000 samples  & $28.9146\pm22.4461$ & $4.0550\pm1.6349$ & $0.2131\pm0.1013$ & 7.4727 \\
& 2000 samples  & $28.8667\pm22.4426$ & $4.0330\pm1.6527$ & $0.2129\pm0.1002$ & 3.6997 \\
& 5000 samples  & $28.6619\pm22.3158$ & $4.0279\pm1.6343$ & $0.2100\pm0.1012$ & 1.4071 \\
& 10000 samples & $28.1517\pm22.2918$ & $3.8678\pm1.6300$ & $0.2071\pm0.1016$ & 0.6501 \\ \cline{2-6} 
\multirow{3}{*}{\makecell[c]{GMM methods\\(Ours)}}
& 1-sigma region & $22.6850\pm16.4856$ & $1.6504\pm0.9148$ & $0.1136\pm0.0628$ & 285.9330 \\
& 2-sigma region & $22.6639\pm16.0608$ & $1.6589\pm0.9099$ & $0.1145\pm0.0622$ & 289.1035 \\
& 3-sigma region & $22.6475\pm15.6722$ & $1.6620\pm0.9072$ & $0.1151\pm0.0624$ & 290.8669 \\
\bottomrule
\end{tabular}
}
\end{table*}

\begin{figure*}[htbp]
\centering
\subfloat[Experimental result of the inter-gender task]{
\includegraphics[width=0.32\linewidth]{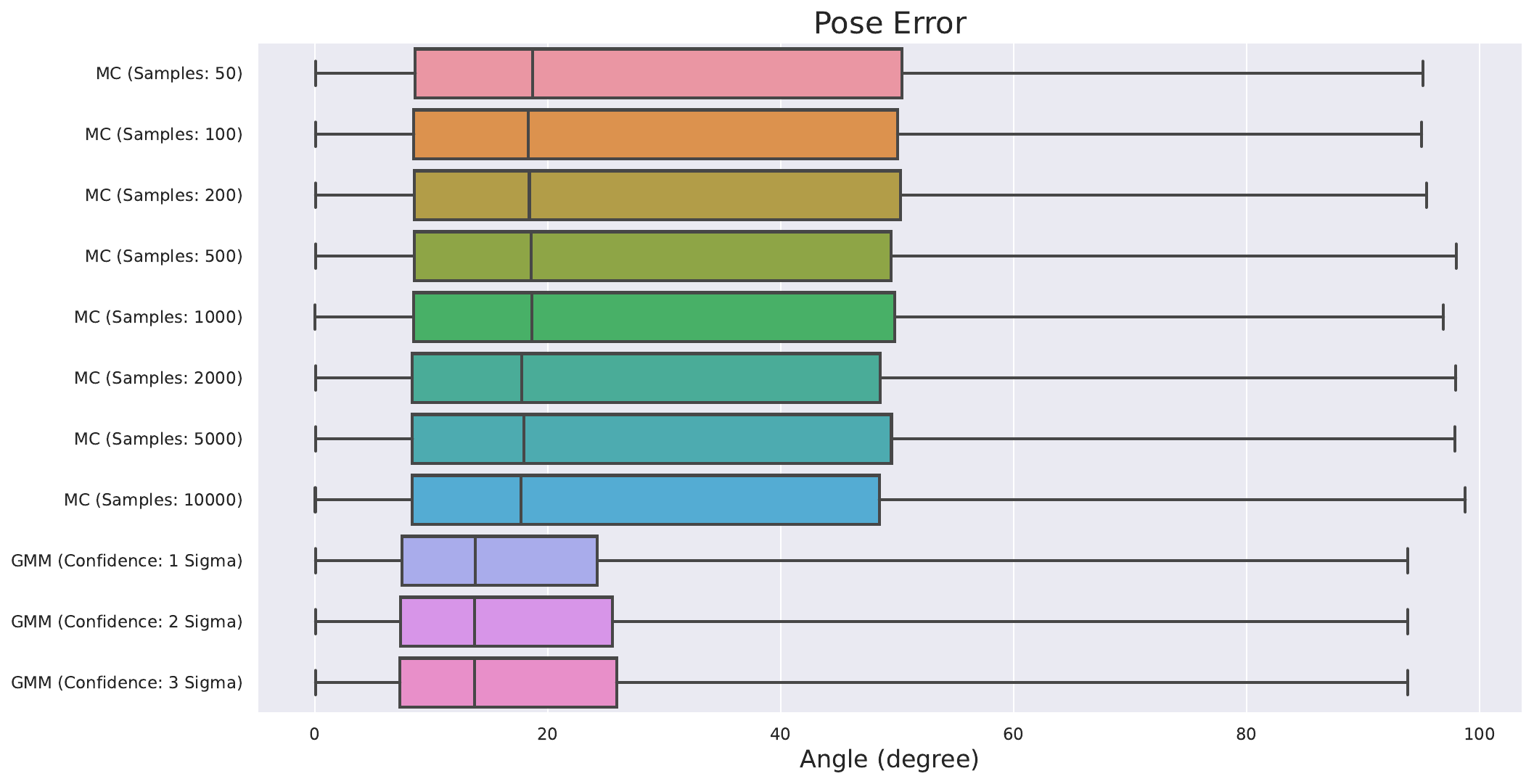}
\includegraphics[width=0.32\linewidth]{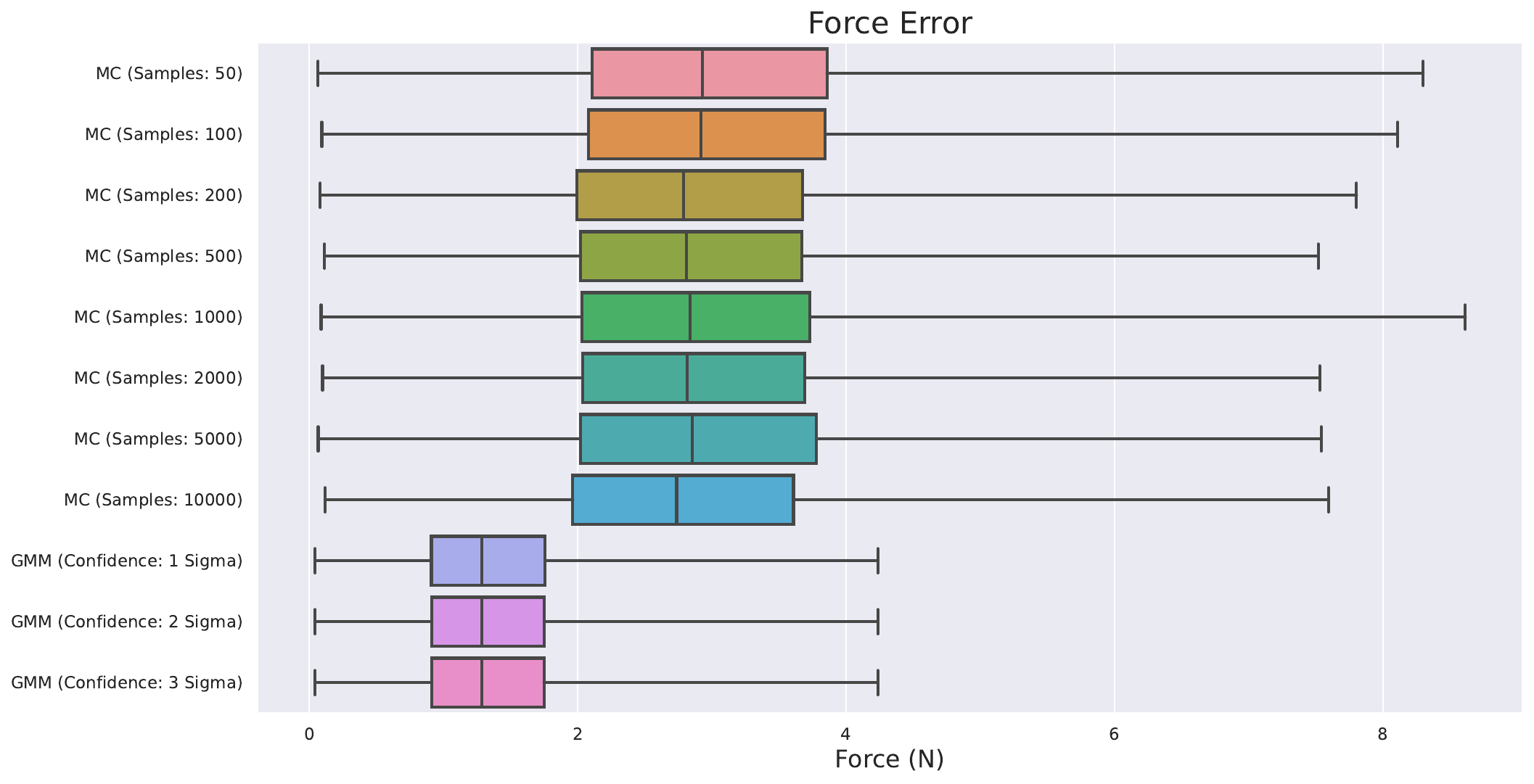}
\includegraphics[width=0.32\linewidth]{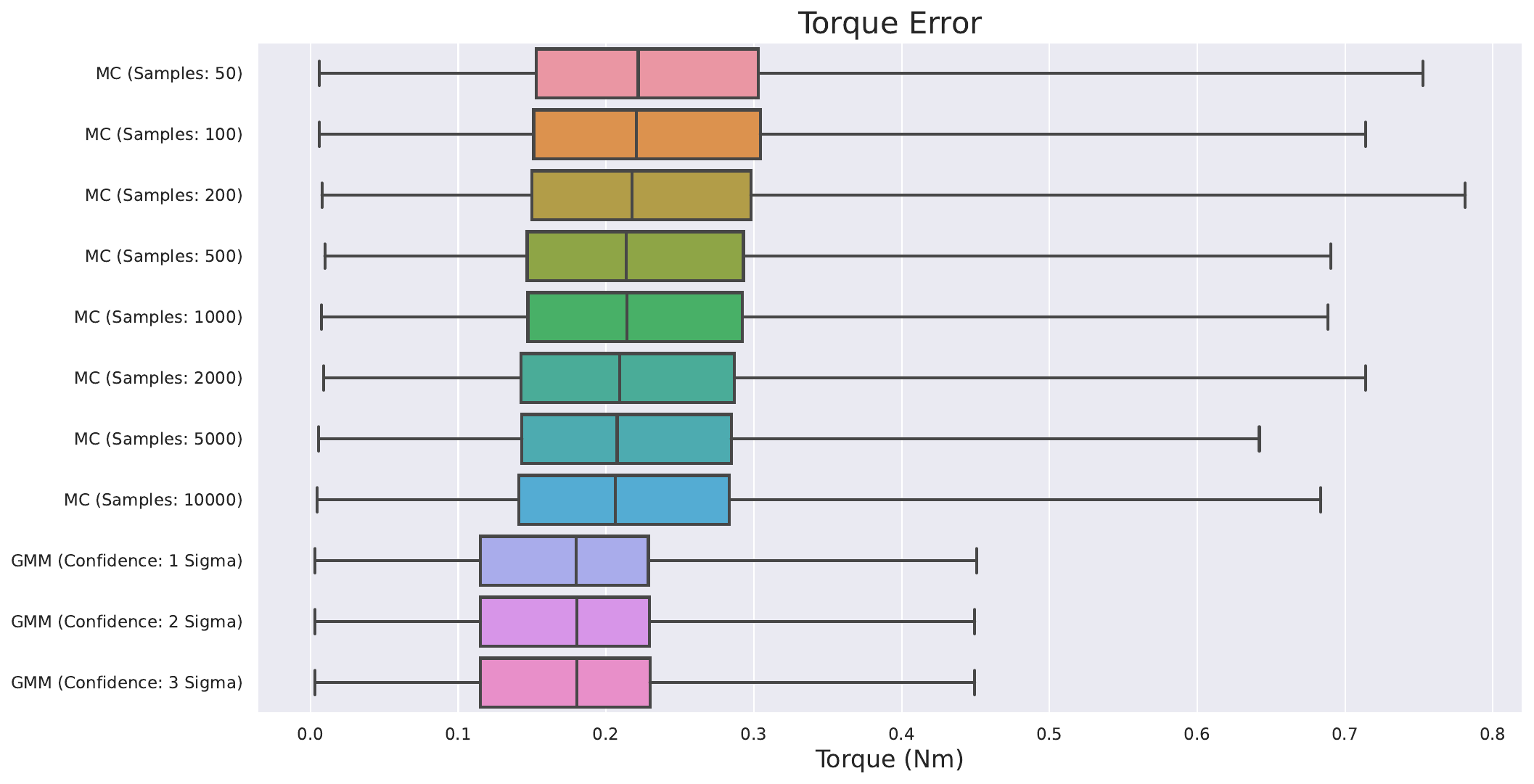}
}
\quad
\subfloat[Experimental result of the inter-age task]{
\includegraphics[width=0.32\linewidth]{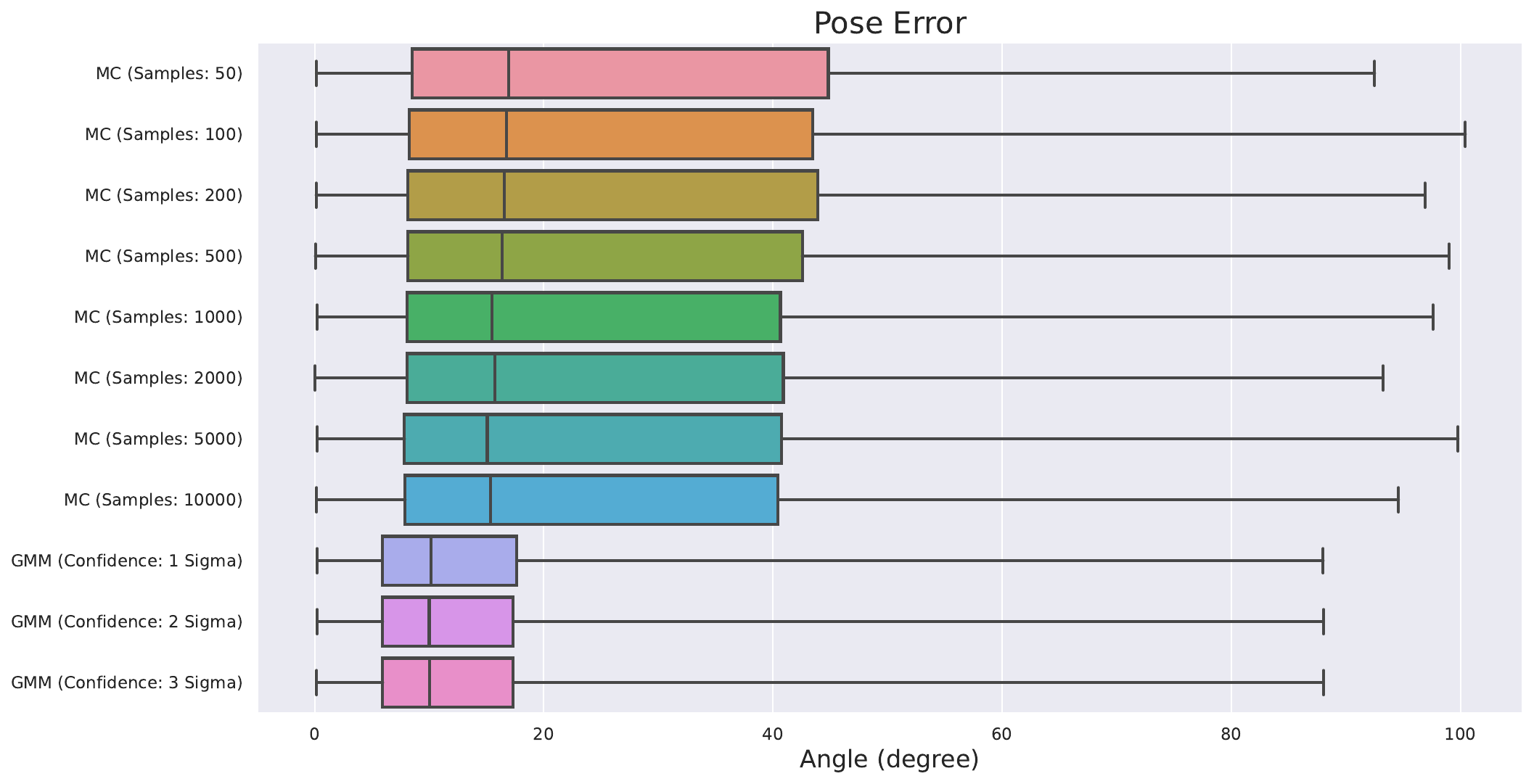}
\includegraphics[width=0.32\linewidth]{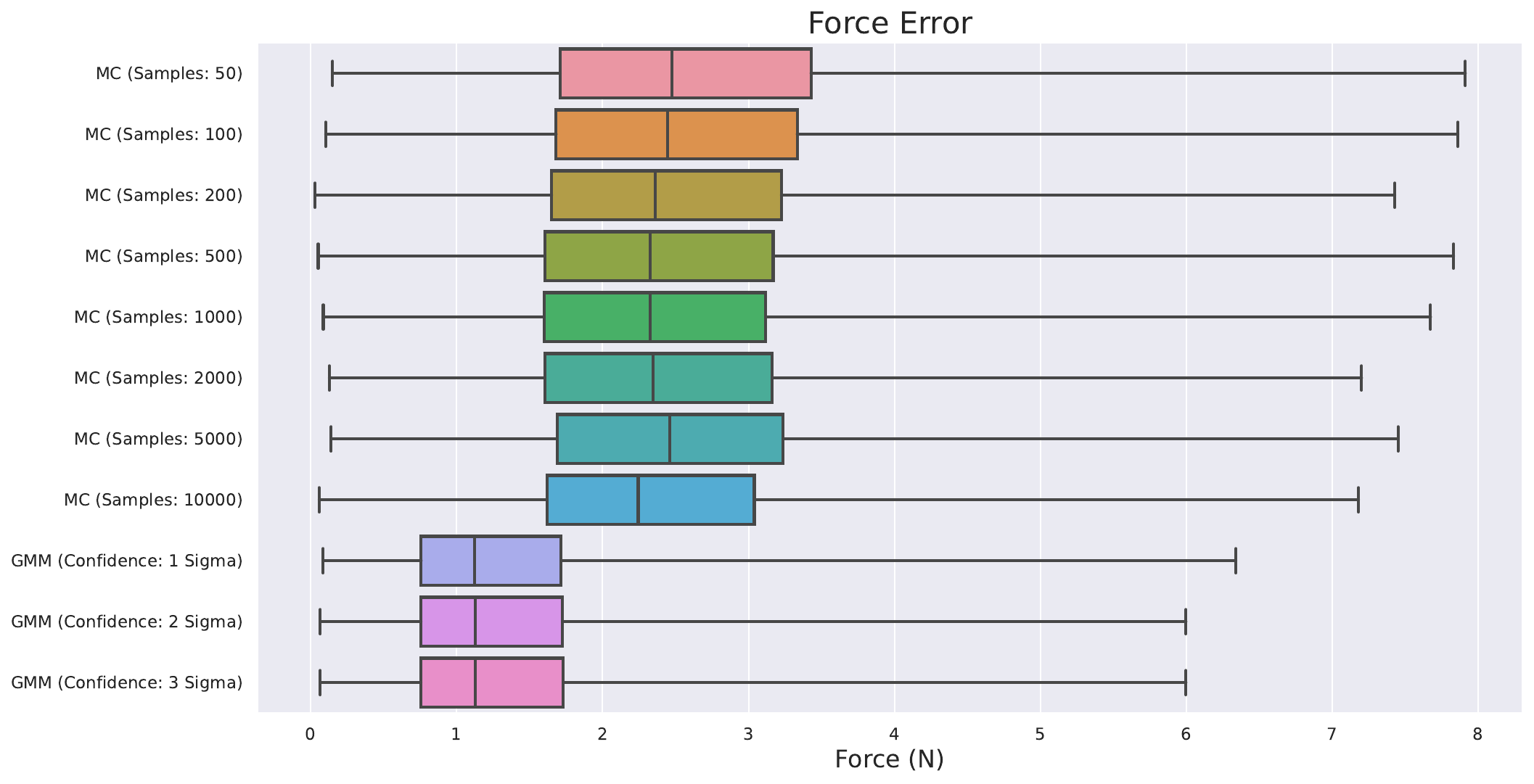}
\includegraphics[width=0.32\linewidth]{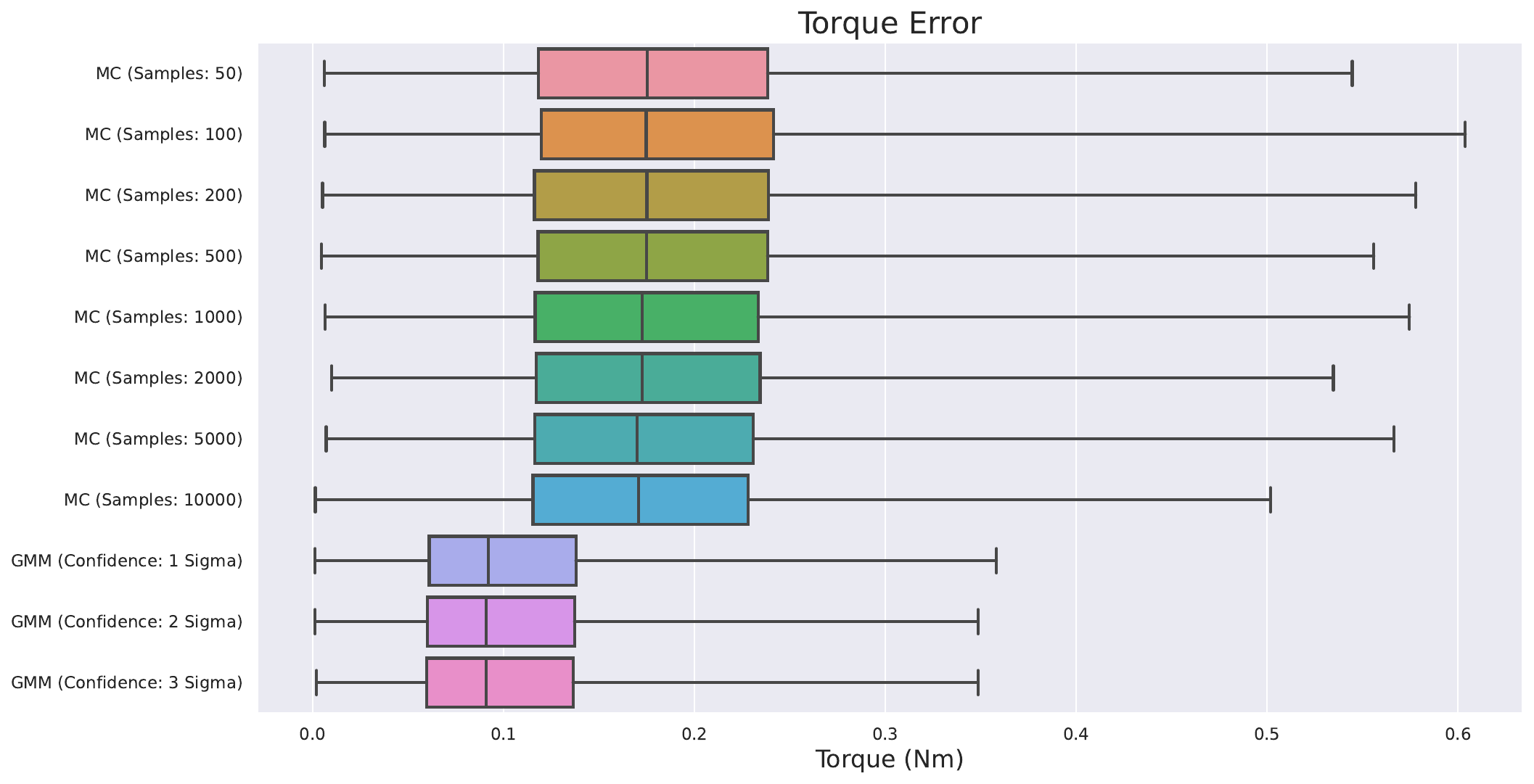}
}
\quad
\subfloat[Experimental result of the inter-BMI task]{
\includegraphics[width=0.32\linewidth]{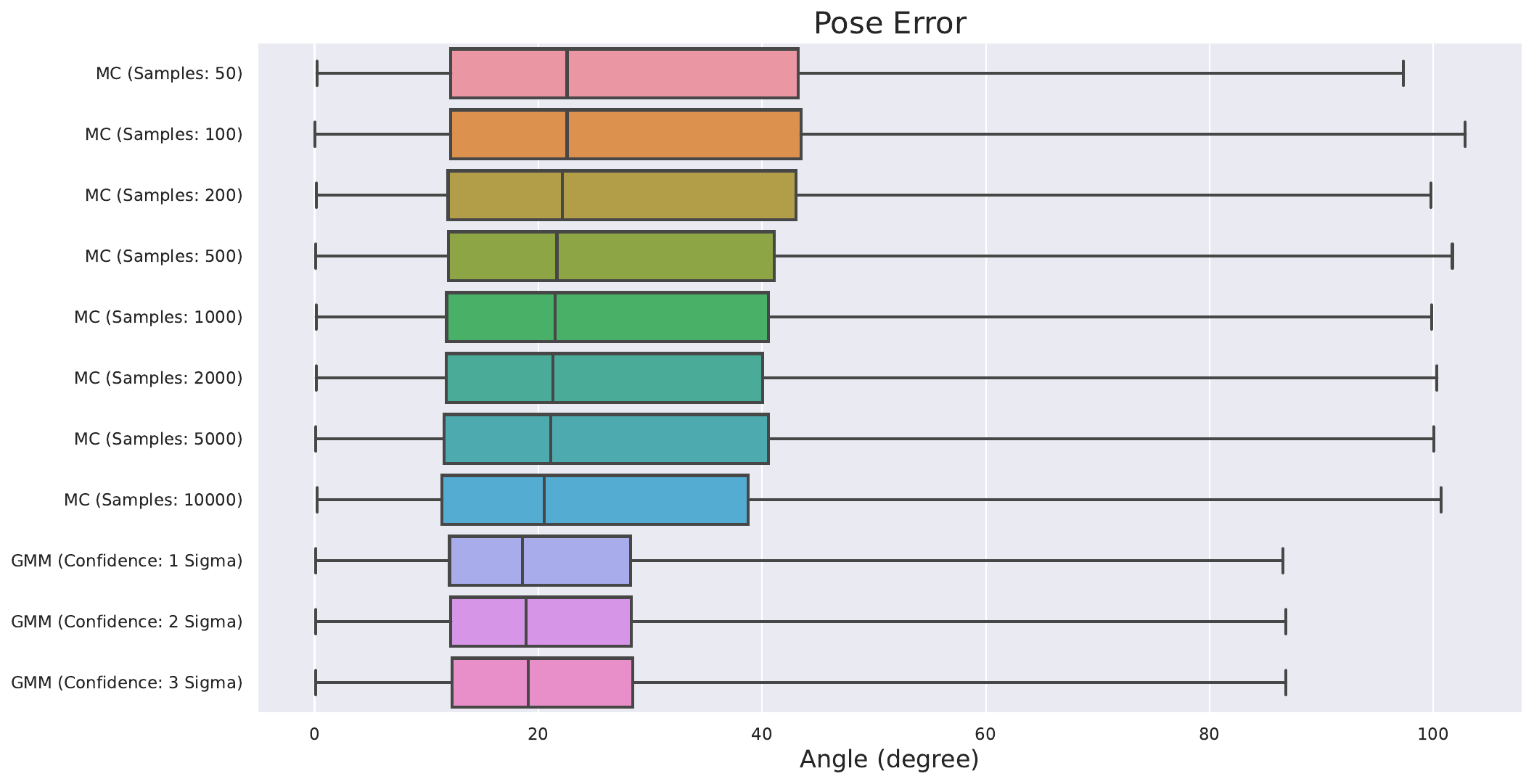}
\includegraphics[width=0.32\linewidth]{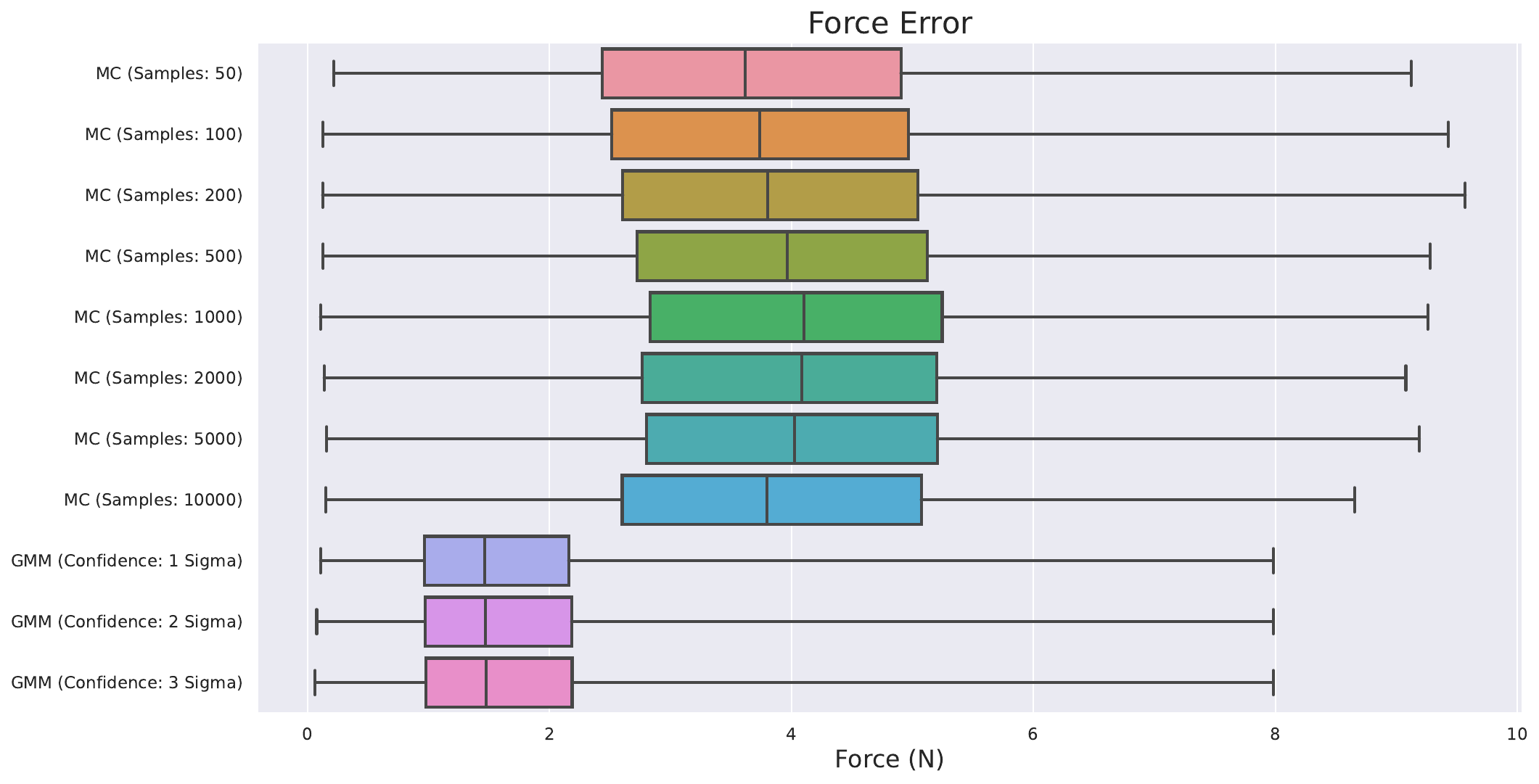}
\includegraphics[width=0.32\linewidth]{figs/3-pe.pdf}
}
\caption{Experimental results of inter-gender, inter-age, inter-BMI tasks.}
\end{figure*}